\documentclass[sigconf, nonacm]{acmart}

\AtBeginDocument{%
  }

\usepackage{microtype}
\usepackage[ruled,linesnumbered]{algorithm2e}
\usepackage{amsmath}
\usepackage{setspace}
\usepackage{multirow}
\usepackage{makecell}
\usepackage{xcolor}

\begin{document}

\title{ResAware: Cross-Environment Website Fingerprinting via Resource-Privileged Distillation}


% ====================================================================
% 1. 范崇儒 (Chongru Fan) - 第一作者
% ====================================================================
\author{Chongru Fan}
\email{Chongrufan@bupt.edu.cn} % 请替换为你的真实邮箱
\affiliation{%
  \institution{Beijing University of Posts and Telecommunications}
  \city{Beijing}
  \country{China}
}
\affiliation{%
  \institution{Zhongguancun Laboratory}
  \city{Beijing}
  \country{China}
}

% ====================================================================
% 2. 王玮 (Wei Wang)
% ====================================================================
\author{Wei Wang}
\affiliation{%
  \institution{Zhongguancun Laboratory}
  \city{Beijing}
  \country{China}
}

% ====================================================================
% 3. 黄文涛 (Wentao Huang)
% ====================================================================
\author{Wentao Huang}
\affiliation{%
  \institution{Beijing University of Posts and Telecommunications}
  \city{Beijing}
  \country{China}
}

% ====================================================================
% 4. 丁振全 (Zhenquan Ding) - 通讯作者
% ====================================================================
\author{Zhenquan Ding}
\authornote{Corresponding author.} % 标注通讯作者
\email{dingzq@zgclab.edu.cn} % 请替换为导师真实邮箱
\affiliation{%
  \institution{Zhongguancun Laboratory}
  \city{Beijing}
  \country{China}
}

% ====================================================================
% 5. 时金桥 (Jinqiao Shi) - 共同通讯作者
% ====================================================================
\author{Jinqiao Shi}
\authornote{Joint corresponding author.} % 标注共同通讯作者
\email{shijinqiao@bupt.edu.cn} % 请替换为导师真实邮箱
\affiliation{%
  \institution{Beijing University of Posts and Telecommunications}
  \city{Beijing}
  \country{China}
}

% ====================================================================
% 6. 崔磊 (Lei Cui)
% ====================================================================
\author{Lei Cui}
\affiliation{%
  \institution{Zhongguancun Laboratory}
  \city{Beijing}
  \country{China}
}

% ====================================================================
% 7. 郝志宇 (Zhiyu Hao)
% ====================================================================
\author{Zhiyu Hao}
\affiliation{%
  \institution{Zhongguancun Laboratory}
  \city{Beijing}
  \country{China}
}

% ====================================================================
% 8. 云晓春 (Xiaochun Yun)
% ====================================================================
\author{Xiaochun Yun}
\affiliation{%
  \institution{Zhongguancun Laboratory}
  \city{Beijing}
  \country{China}
}

\renewcommand{\shortauthors}{C. Fan et al.}

\begin{abstract}

While Website Fingerprinting (WF) attacks achieve high accuracy in controlled laboratory settings, they often degrade substantially in real-world environments due to spatio-temporal drift, browser heterogeneity, proxy obfuscation and etc. This limitation stems from their sole reliance on low-level traffic features that are noisy and highly sensitive to environmental perturbations. To address this problem, we propose \textbf{ResAware}, a cross-environment resource-aware distillation framework under a \textit{training-rich/inference-poor} asymmetric setting. Specifically, ResAware trains a teacher model on resource-level features, and then distills the resulting privileged knowledge into a student model through heterogeneous knowledge distillation. At deployment time, the student model performs inference using only encrypted traffic, incurring zero additional cost. We evaluate ResAware on a large-scale dataset collected over five months from six globally distributed vantage points, comprising more than $160{,}000$ paired samples. The results show that ResAware significantly enhances the cross-environment robustness of diverse WF baselines. Under a 150-day temporal drift, for example, ResAware improves the F1-score of Var-CNN from $72.77\%$ to $81.49\%$ and the open-world $TPR@1\%FPR$ from $22.40\%$ to $27.20\%$. Our results demonstrate that resource-level supervision improves WF robustness without expanding online observation capabilities.

\end{abstract}



%     \includegraphics[width=\textwidth]{sampleteaser}
%     \caption{Seattle Mariners at Spring Training, 2010.}
%     \Description{Enjoying the baseball game from the third-base
%         seats. Ichiro Suzuki preparing to bat.}
%     \label{fig:teaser}


\maketitle

\section{Introduction}

With the widespread adoption of HTTPS and related encryption protocols, the contents of web traffic are now largely hidden from direct inspection~\cite{ietf-tls-esni-25, rfc8484}. However, encryption does not eliminate side-channel leakage: observable traffic patterns, such as packet length, direction, and timing, can still reveal sensitive information about user activities. Website Fingerprinting (WF) exploits such leakage to infer visited websites from encrypted traffic traces~\cite{fingerprinting2002hintz, hayes2016kFingerprinting, wang2014effectiveAttacks}, making it an important privacy threat to encrypted web communications.

\begin{figure}[t]
    \centering
    \includegraphics[width=1\linewidth]{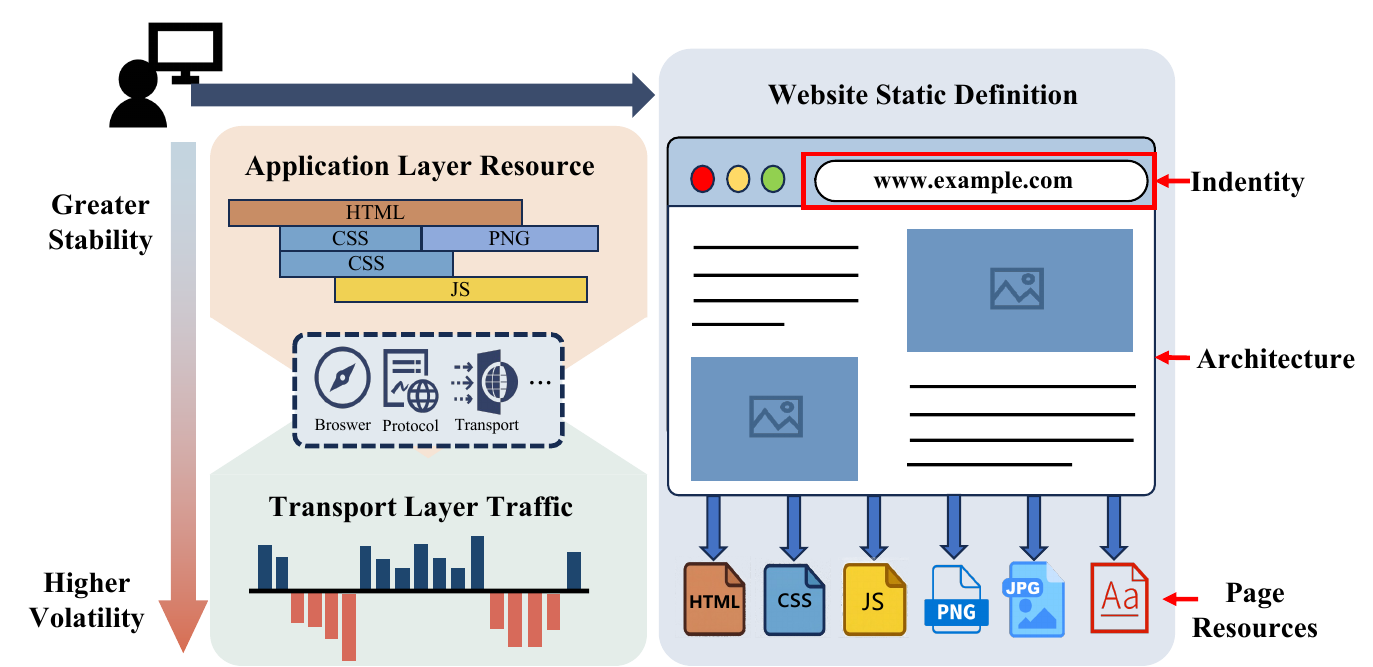}
    \caption{A website's identity is reflected in its architecture and resource loading patterns. A browsing instance can be viewed as a sequence of resource deliveries, which, after being shaped by environmental noise, appears as the observable network traffic.}
    \Description{A diagram illustrating that a website's static resource dependency topology generates a stable application-layer resource sequence, which is then mapped to an encrypted traffic trace. ResAware uses the resource sequence as privileged offline supervision while keeping the online attacker restricted to encrypted traffic only.}
    \label{fig:insight}
\end{figure}

Although deep learning-based WF models have achieved strong performance in closed, IID experimental settings~\cite{sirinam2018deepFingerprinting, bhat2019varcnn, rahman2020tiktok, deng2025countmamba}, a substantial deployment gap remains between laboratory scenarios and real-world network environments. In practice, traffic features are highly susceptible to temporal evolution, geographic variation, and obfuscated proxy protocol conversion, all of which induce significant distribution shifts~\cite{cherubin2022online, deng2025singlePerspective, li2025crossEnvironmental, shusterman2026conceptDrift}. When a model is trained in one environment and evaluated in another with substantial feature discrepancies, the accuracy of mainstream WF models deteriorates markedly. This degradation suggests that existing models rely heavily on transient, environment-specific network artifacts and therefore generalize poorly across environments.

Prior work on mitigating this problem generally follows two trajectories. The first seeks more robust traffic representations through manual feature engineering, data augmentation, or contrastive learning~\cite{shen2023robustTrafficRepresentation, bahramali2023realistic, xie2024contrastive}. The second adopts domain adaptation, such as few-shot fine-tuning or inference-time calibration on unlabeled target traffic~\cite{sirinam2019triplet, deng2026proteus, zhang2023unsupervised}. Despite these efforts, existing methods remain confined to a \textit{traffic-only} observational perspective: both training and inference rely exclusively on signals derived from encrypted traffic. Such signals are readily distorted by network variation, browser scheduling, and protocol encapsulation. As a result, current methods attempt to recover website identities from unstable observations, while overlooking the deterministic application-layer resources that give rise to these traffic patterns.

As illustrated in Figure~\ref{fig:insight}, our key insight is that a website's identity determines its application-layer resource composition and dependency patterns~\cite{demystifying2013xiao, polaris2016ravi}. During page loading, the resulting resource sequence reflects relatively stable website-specific loading logic. By contrast, the observed network traffic is only a noisy projection of this process, shaped by substantial stochasticity and environmental variation~\cite{critical2014juarez}. Recent resource-aware WF studies suggest that resource-level information enjoys a natural robustness advantage in cross-environment settings~\cite{cheng2025holmesWatson, gao2025mrcgcn, cheng2025star}. However, obtaining such information in practice typically requires traffic decryption or end-host compromise, both of which exceed the capabilities of a standard passive eavesdropper~\cite{Panchenko2016WebsiteFA}.

To exploit resource-level stability without expanding the online attack surface, we formalize an asymmetric threat setting termed \textit{training-rich/inference-poor}. During offline fingerprint database construction, the attacker can collect both encrypted traffic and resource-level information using controlled crawlers; during online inference, however, the attacker remains a standard passive eavesdropper limited to encrypted traffic alone. Under this setting, resource-level information is available during training but unavailable during online inference, which naturally qualifies it as Privileged Information~\cite{vapnik2015privileged}: an auxiliary supervisory signal that guides learning without being available as an input at inference time.

Motivated by the Learning Using Privileged Information (LUPI) paradigm~\cite{vapnik2015privileged, lopezpaz2016unifying}, we propose ResAware, a resource-aware distillation framework for cross-environment WF under the \textit{training-rich/inference-poor} setting. Using paired traffic-resource samples, ResAware trains a resource-side teacher model on resource-level features and distills the resulting privileged knowledge into a student model through cross-modal knowledge distillation~\cite{hinton2015distilling, lopezpaz2016unifying}. The student model operates on encrypted traffic alone. At deployment time, the resource sequences and the teacher model are removed, leaving a standard traffic-only WF classifier. In this way, ResAware improves robustness without strengthening the online attacker beyond the standard passive threat model.

We evaluate ResAware under multidimensional distribution shifts, including temporal, spatial, browser, and proxy variations, on a large-scale dataset spanning five months across six globally distributed nodes and more than 160{,}000 samples. The results show that ResAware robustly improves the cross-environment generalization of mainstream WF baselines with zero additional inference cost. Moreover, ResAware is orthogonally complementary to existing target-domain adaptation techniques~\cite{sirinam2019triplet, deng2026proteus, zhang2023unsupervised}.

The main contributions of this paper are as follows:
\begin{itemize}

    \item \textbf{Asymmetric Threat Model Formalization.} We introduce and formalize a \textit{training-rich/inference-poor} asymmetric setting for WF. Under this setting, application-layer resource information is available only during training and naturally serves as privileged information, improving the robustness of \textit{traffic-only} WF models while preserving the standard passive eavesdropper assumption.

    \item \textbf{Cross-Modal Distillation Framework.} We propose ResAware, a cross-modal knowledge distillation framework for cross-environment WF. ResAware trains a resource-side teacher model on resource-level features and distills the resulting privileged knowledge into a \textit{traffic-only} student model, injecting stable resource-side supervision without requiring resource access at inference time.

    \item \textbf{Plug-and-Play Integration with Zero Inference Overhead.} ResAware can be incorporated into existing WF models through the training objective alone, without modifying backbone architectures. At deployment time, it operates directly on encrypted traffic with zero additional inference overhead and remains complementary to existing domain adaptation techniques.

    \item \textbf{Large-Scale Benchmark and Evaluation.} We construct a large-scale paired traffic-resource dataset spanning multidimensional cross-environment scenarios. On this benchmark, ResAware robustly improves the robustness of mainstream WF baselines. Under a 150-day temporal drift, it raises the F1-score of Var-CNN from 72.77\% to 81.49\% and improves the open-world TPR from 22.40\% to 27.20\% at 1\% FPR. Under the more challenging obfuscated proxy drift setting, it further delivers absolute F1-score gains of 8.96\% and 3.88\% for Var-CNN and RF, respectively.

\end{itemize}

\section{Threat Model}

\begin{figure}[t]
    \centering
    \includegraphics[width=0.95\linewidth]{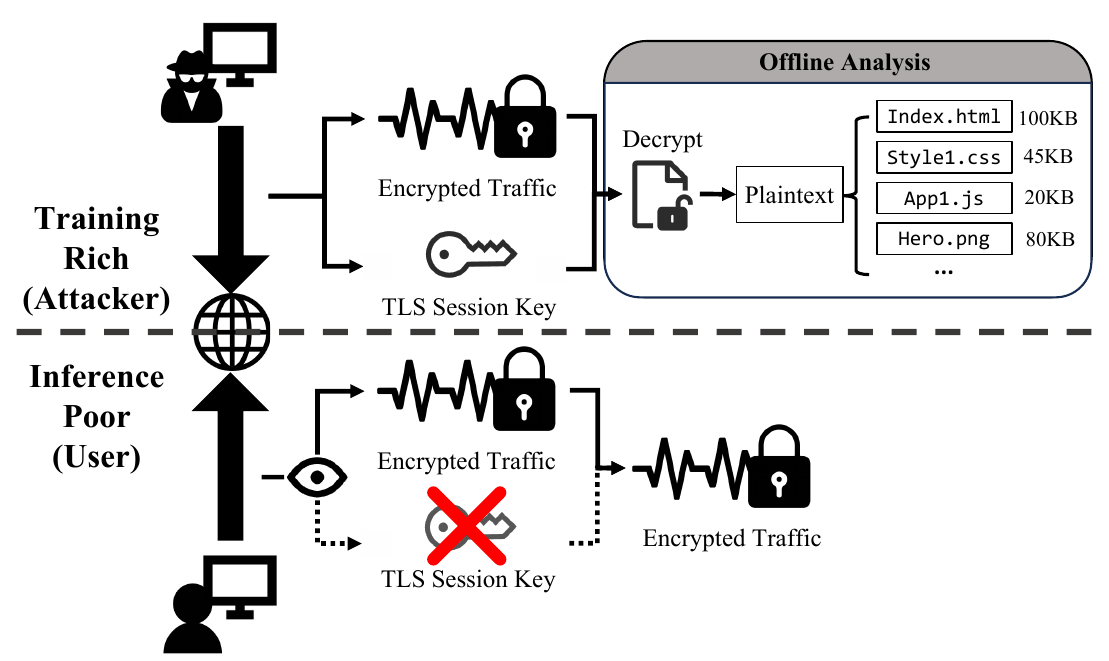}
    \caption{The \textbf{training-rich / inference-poor} asymmetric threat model. During offline construction, the attacker collects paired traffic and resource sequences via TLS key logging; at online inference, only encrypted traffic is observable.} 
    \Description{A two-phase diagram showing the offline construction phase where the attacker collects paired traffic and resource sequences via TLS key logging, and the online inference phase where the attacker is a passive eavesdropper observing only encrypted packet-level traffic.}
    \label{fig:threat_model}
\end{figure}

As depicted in Figure~\ref{fig:threat_model}, we formalize an asymmetric threat model for cross-environment WF, termed \textit{training-rich/inference-poor}. The asymmetry lies in the fact that resource-level information is available during offline training but unavailable during online inference.

\textbf{Offline Construction Phase.} As shown in the upper portion of Figure~\ref{fig:threat_model}, the attacker deploys instrumented crawlers in a controlled environment to visit websites and collect both encrypted traffic and the corresponding TLS key logs. These key logs are generated solely by attacker-controlled crawlers during offline data collection and are not available from the victim during online inference. This enables offline parsing of encrypted traffic and extraction of high-fidelity application-layer resource information, which is used solely as privileged supervision during training.

\textbf{Online Inference Phase.} The lower portion of Figure~\ref{fig:threat_model} depicts the online attacker as a client-side path observer, such as a local Autonomous System (AS), an Internet Service Provider (ISP), or a malicious local router. The attacker can only passively observe packet-level characteristics of the victim's encrypted connections, including packet direction, length, and timing. The attacker cannot decrypt payloads, inject, modify, delay, or drop packets, and has no control over the victim's endpoint. We further exclude side-channel metadata that could directly reveal the target website, including DNS queries, TLS SNI fields, certificate contents, HTTP Host headers, IP-to-domain mappings, and browser-side API fingerprints. Consistent with mainstream page-load-level WF research~\cite{cheng2025star,sirinam2018deepFingerprinting,huang2023efficient}, we assume that each test sample corresponds to a single isolated page-load event.

\section{Motivating Analysis: Stability and Transferability of Resource-Level Features}

To motivate ResAware, we examine two fundamental questions. \textbf{(1) Resource-Level Feature Robustness:} Are resource-level features stable and discriminative under cross-environment distribution shifts? \textbf{(2) Knowledge Transferability:} Can knowledge derived from offline resource-level information be effectively distilled into a \textit{traffic-only} student model?

\subsection{Are Resource-Level Features Stable and Discriminative Across Environments?}

Modern web page loading follows a structured process shaped by HTML, CSS, JavaScript, and asynchronous resource fetching~\cite{rfc9110}. While dynamic updates, advertisement injections, and A/B testing may introduce localized variation, the overall resource loading sequence of a website typically retains stable macroscopic patterns across visits~\cite{li2023robust,panchenko2016website}. In contrast, low-level traffic features, such as packet length, direction, and burst intervals, are heavily affected by transport- and network-layer dynamics, including congestion, routing changes, and TCP control behavior. This suggests that resource loading sequences may provide a more stable basis for cross-environment identification than low-level traffic features.

To validate this hypothesis, we conduct an empirical analysis along two dimensions. First, we quantify the cross-environment drift and class separability of resource representations in feature space (\textbf{Finding 1}). Second, we evaluate whether this relative stability yields more robust classification performance under cross-environment distribution shifts (\textbf{Finding 2}).

\begin{figure}[t]
    \centering
    \includegraphics[width=\linewidth]{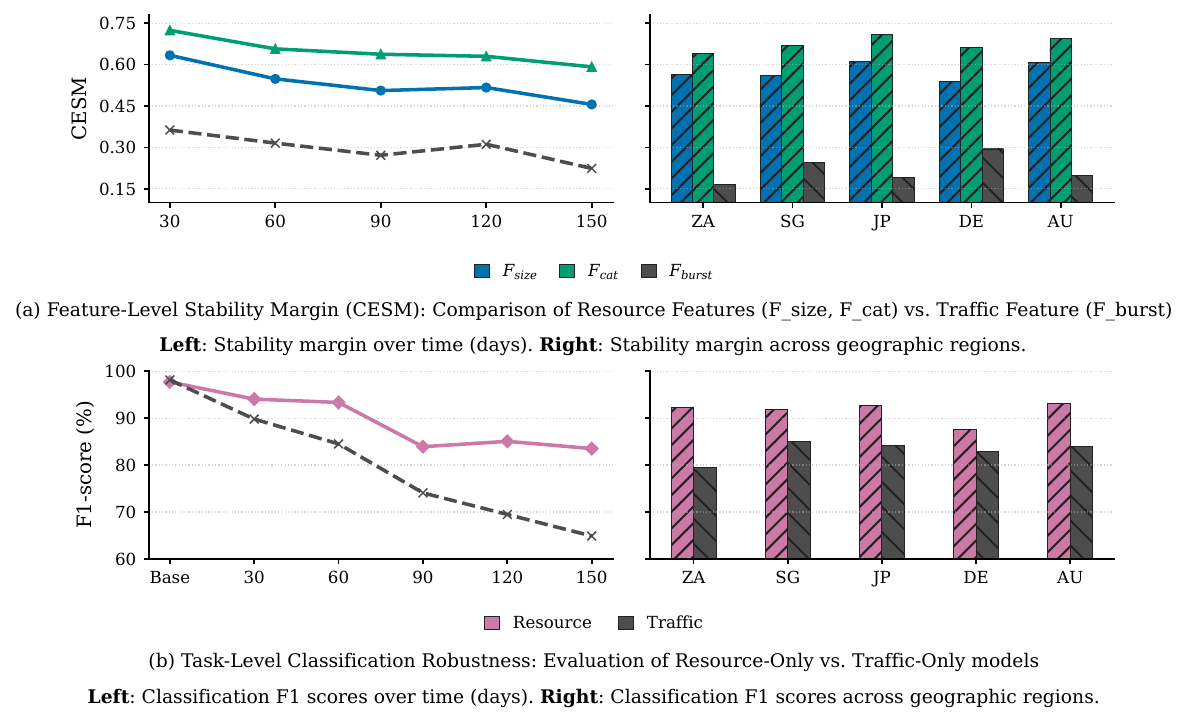}
    \caption{(a) CESM comparison between resource features ($F_\mathrm{cat}$, $F_\mathrm{size}$) and traffic bursts ($F_\mathrm{burst}$) under spatial and temporal drift; (b) classification F1-scores decay for a resource-only vs.\ traffic-only model.}
    \label{fig:motivation_2x2}
\end{figure}

\noindent \textbf{Finding 1: Compared with traffic features, resource-level features exhibit substantially stronger intra-class stability and inter-class separability across environments.}

For each page load, we extract two resource sequences ordered by request initiation time. The first, $F_{size}$, records the log-scaled payload size of each fetched resource. The second, $F_{cat}$, represents resource categories as one-hot vectors. As a traffic-side baseline, we derive $F_{burst}$ from encrypted traces by grouping contiguous packets traveling in the same direction and representing each burst by its signed log-scaled size.

We perform cross-regional and cross-temporal measurements on 100 monitored websites. Because these sequences have variable length and may exhibit local alignment shifts under cross-environment loading conditions~\cite{li2025crossEnvironmental}, we use Normalized Dynamic Time Warping (nDTW)~\cite{berndt1994dtw} to measure distances between site prototypes. For a feature $F$, we define the same-site cross-environment drift ($\Delta_{\text{same}}$) and the different-site cross-environment distance ($\Delta_{\text{diff}}$) between source environment $s$ and target environment $t$ as follows:

\begin{equation}
\begin{gathered}
    \Delta_{\text{same}}^{F} = \frac{1}{N} \sum_{i=1}^{N} d(p_{i,s}^{F}, p_{i,t}^{F}), \\
    \Delta_{\text{diff}}^{F} = \frac{1}{N(N-1)} \sum_{i \neq j} d(p_{i,s}^{F}, p_{j,t}^{F}),
\end{gathered}
\end{equation}

To jointly evaluate whether features retain discriminative power while suppressing environmental noise, we introduce the Cross-Environment Stability Margin (CESM):

\begin{equation}
\text{CESM}_{F}(s,t) = 1 - \Delta_{\text{same}}^{F} / \Delta_{\text{diff}}^{F}.
\end{equation}

A higher CESM indicates that cross-environment intra-site variation remains much smaller than inter-site discrepancy, and therefore reflects stronger robustness to environmental noise. As shown in Figure~\ref{fig:motivation_2x2}(a), both $F_{size}$ and $F_{cat}$ exhibit substantially stronger cross-environment stability than $F_{burst}$. In cross-regional experiments spanning five geographic regions, $F_{cat}$ and $F_{size}$ achieve average CESM values of 0.675 and 0.577, respectively, compared with 0.218 for $F_{burst}$, corresponding to gains of 3.09$\times$ and 2.65$\times$ over $F_{burst}$. The same trend persists under temporal drift: after 150 days, $F_{cat}$ (CESM = 0.592) and $F_{size}$ (CESM = 0.456) remain well above $F_{burst}$ (CESM = 0.224), yielding 2.64$\times$ and 2.04$\times$ larger margins, respectively. These results show that resource-level features are inherently more robust to environmental variation and preserve stronger discriminative structure across diverse deployment conditions.

\noindent \textbf{Finding 2: The stability advantage of resource-level features yields stronger task-level robustness.}
To examine whether the feature-space advantage carries over to downstream classification robustness, we control all other factors and compare two classifiers built on the same Deep Fingerprinting (DF)~\cite{sirinam2018deepFingerprinting} architecture: a \textit{Resource-Only} model, which takes $F_{size}$ and $F_{cat}$ as input, and a \textit{Traffic-Only} model, which takes packet-level traffic sequences as input.

As shown in Figure~\ref{fig:motivation_2x2}(b), the \textit{Resource-Only} model degrades much more slowly under environmental drift. Under temporal drift, both models initially achieve near-perfect source-domain F1-scores. However, after 150 days, the \textit{Traffic-Only} model drops by 33.30 percentage points to 64.85\%, whereas the \textit{Resource-Only} model declines by only 14.22 points and still maintains 83.50\%. In cross-regional evaluation, the \textit{Resource-Only} model achieves an average F1-score of 91.49\% across all target regions, outperforming the \textit{traffic-only} model by 8.35 percentage points on average.

\noindent \textbf{Takeaway.} These empirical results show that resource-level features are substantially more stable and robust to environmental noise than low-level traffic features under cross-environment shifts. Yet such robust resource-side signals are unavailable to a passive eavesdropper at deployment time. This gap motivates the core design of ResAware: the key challenge is not to seek stronger features from online observations alone, but to transfer resource-side robustness to a \textit{traffic-only} classifier through offline cross-modal supervision.

\subsection{Can Resource-Side Robustness Be Transferred to Traffic-Only Models?}

This leads to the central methodological question behind ResAware: can the stability available only through privileged supervision during training be transferred across modalities, and if so, how can it improve a classifier that must rely solely on low-level traffic at deployment time?

Our answer is yes, but not by assuming that resource sequences can be faithfully reconstructed from encrypted traffic. In modern web communications, concurrent browser scheduling, HTTP/2/3 multiplexing, transport-layer dynamics, and network latency variation collectively entangle multiple object-level requests within a continuous packet stream. Recovering precise object boundaries from encrypted traffic is therefore highly ill-posed in practice. Building stability transfer on such packet-to-object reconstruction would not only be impractical, but would also force the model to depend on fragile local alignment assumptions.

Instead, ResAware follows a more robust transfer pathway groun-ded in the Learning Using Privileged Information (LUPI) paradigm~\cite{vapnik2015privileged, lopezpaz2016unifying} and generalized knowledge distillation. The resource view, available only during training, does not need to be reconstructed at inference time. As long as it provides a cleaner inductive signal than encrypted traffic alone, it can reshape the decision boundaries of a single-modality classifier through a teacher-student framework. In cross-environment WF, resource-side privileged supervision fits this paradigm particularly well.

Under standard hard-label supervision with Empirical Risk Minimization (ERM), a single-modality model can easily fall into \textit{shortcut learning}, relying on spurious yet separable packet-level cues in the source environment and thus forming brittle decision boundaries that fail under distribution shift.

ResAware mitigates this problem by introducing a structural prior from the resource modality through soft-target supervision. This prior captures class-level similarity relationships at the application layer and guides the \textit{traffic-only} model toward representations that better reflect intrinsic website identity, rather than transient environment-specific traffic patterns. What is transferred is not the raw resource sequence itself, but the class-level relational knowledge encoded in the resource modality. The appropriate role of the resource modality is therefore not as an auxiliary runtime input, but as a source of privileged supervision during training. How effectively this knowledge is inherited by the student, and how it affects decision-boundary calibration under long-term drift, are quantitatively analyzed in \S\ref{sec:mechanism_analysis}.

\section{ResAware Overview and Design}
\label{sec:design}

\begin{figure*}[t]
    \centering
    \includegraphics[width=0.92\textwidth]{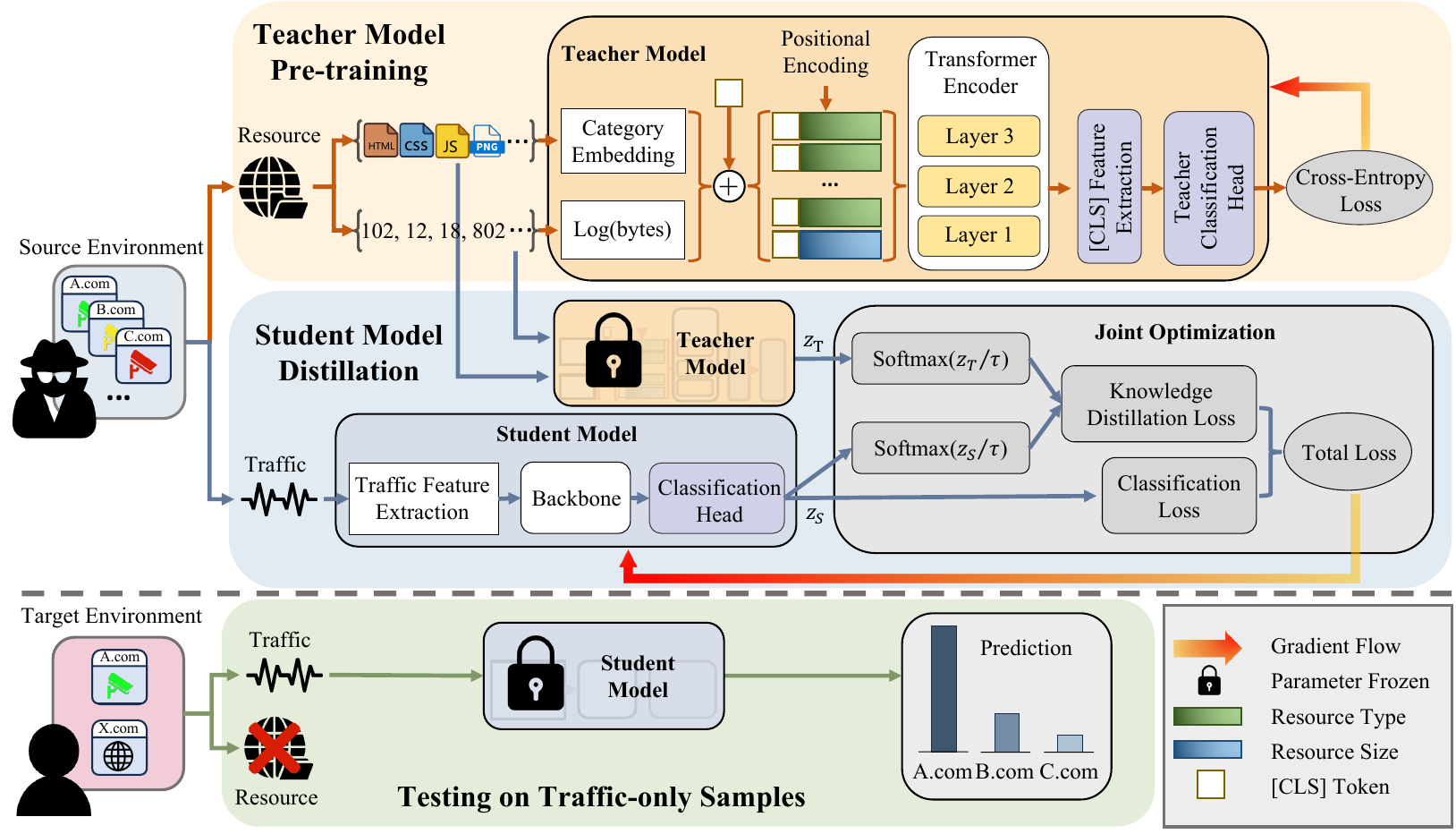}
    \caption{Overview of the ResAware framework. Offline training first trains a resource-only teacher, then distills its knowledge into a traffic-only student; all resource-side components are discarded before online deployment.}
    \Description{A pipeline diagram of ResAware divided into two phases: offline training with resource extraction, resource teacher training, and cross-modal knowledge distillation to the traffic-only student; and online inference where only the traffic student is deployed with no resource access.}
    \label{fig:resaware_framework_pipeline}
\end{figure*}

ResAware is a training-time privileged knowledge distillation framework for cross-environment WF. Its core idea is straightforward: during the offline construction phase, a \textit{resource-only teacher model} is trained on resource loading sequences collected under controlled conditions, and the resulting resource-side knowledge is transferred to a \textit{traffic-only student model} through heterogeneous cross-modal distillation. In this way, ResAware introduces stable resource-side supervision into the student model without requiring resource access at inference time.

The framework is designed under three constraints:
\begin{itemize}
    \item \textbf{Privileged Information Isolation.} High-fidelity resource features are available only during offline training and remain inaccessible during online inference.
    \item \textbf{Zero Online Overhead and Interface Compatibility.} At deployment time, the framework must preserve the standard traffic-only observation interface of a passive WF attacker, without introducing additional online cost.
    \item \textbf{Plug-and-Play Integration.} ResAware makes no assumptions about the underlying WF backbone. It can be instantiated on top of any existing WF model following a standard end-to-end classification pipeline.
\end{itemize}

\subsection{Design Principle: Privileged Resource Distillation}

As illustrated in Figure~\ref{fig:resaware_framework_pipeline}, ResAware operates in two strictly separated phases: offline training and online inference. During offline training, the framework has access to paired samples $(x, x^*, y)$, where $x$ denotes the encrypted traffic trace of a single page load, $x^*$ denotes the corresponding resource loading sequence extracted under controlled conditions, and $y$ is the ground-truth website label.

The distillation pipeline proceeds in three steps. First, a \textit{resource-only teacher model} is trained in the offline source environment to capture stable resource-side patterns that are less affected by network noise. Second, the teacher model is frozen, and its soft target outputs are used to supervise a \textit{traffic-only student model}, transferring resource-side relational knowledge into the traffic feature space. Third, all resource-side components are discarded before deployment. The complete training and deployment procedure is summarized in Algorithm~\ref{alg:resaware} (Appendix~\ref{sec:appendix_c}).

\subsection{Training the Resource-Only Teacher}
\label{sec:training_teacher_model}

In the controlled source environment, ResAware constructs page-load-level paired triplets $(x, x^*, y)$, where $x$ denotes the encrypted traffic trace, $x^*$ denotes the corresponding resource loading sequence, and $y$ is the website label. This requires only page-load-level correspondence between $x$ and $x^*$, avoiding any need for fragile packet-to-object reconstruction.

To improve robustness under cross-environment extraction, we order resource events by request \textit{initiation time}, as triggered by the browser engine, rather than by response completion order. Initiation order more directly reflects the parsing progress of the root document and the triggering logic of resource dependencies, whereas completion order is far more sensitive to network latency, congestion control, and HTTP multiplexing. Using initiation order therefore decouples the resource sequence from transport-side timing variation.

To convert a variable-length resource loading sequence into a fixed-size model input, we represent each page load as a sequence of $N$ resource events:

\begin{equation}
    Z = \{(c_i, \tilde{s}_i)\}_{i=1}^{N}
\end{equation}
where $c_i$ is the resource category ID and $\tilde{s}_i$ is the log-scaled payload size. Unless otherwise specified, we set $N = 200$ for truncation and padding. The sequence captures two feature channels:

\begin{itemize}
    \item \textbf{Categorical Channel.} Based on the \texttt{Content-Type} field in HTTP response headers, each resource is mapped into one of nine categories: HTML, Tiny Image, Regular Image, CSS, JS, Font, JSON/API, Document, and Unknown. Image resources are further divided by payload size: those smaller than 5\,KB are classified as Tiny Image, while those of 5\,KB or larger are classified as Regular Image. This taxonomy captures the major resource types that commonly appear in modern web pages.
    \item \textbf{Size Channel.} We use the byte size of each resource as a continuous feature. To reduce the influence of large resources on optimization, the size of the $i$-th resource is log-scaled to $\tilde{s}_i$ before being fed into the model.
\end{itemize}

We deliberately discard absolute request timestamps and preserve temporal structure only through event order, encoded by positional embeddings. This design prevents the teacher from overfitting to source-environment-specific latency patterns. The resulting fixed-length sequence is then used to train a teacher model $T(\cdot)$ based on a Transformer encoder~\cite{Vaswani2017Transformer}, whose parameters are frozen after supervised training on source-domain hard labels.

\subsection{Distilling Resource Knowledge into a Traffic-Only Student}

During cross-modal knowledge distillation, the student model $S(\cdot)$ receives only encrypted traffic as input. ResAware is agnostic to the student's input representation and can be instantiated on top of any compatible WF backbone. The original input format—whether packet length sequences, burst sequences, or Traffic Aggregation Matrices (TAM)—remains unchanged. At each forward pass, the frozen teacher model produces logits $z_T = T(x^*)$, while the student model produces logits $z_S = S(x)$. The student parameters $\theta_S$ are optimized using two loss terms:

\textbf{Classification Loss.} To preserve the student's discriminative accuracy on the traffic modality, we compute the cross-entropy loss against ground-truth labels:
\begin{equation}
    \mathcal{L}_{cls} = - \sum_{c=1}^{C} y_c \log\left(\frac{\exp(z_{S,c})}{\sum_{j=1}^C \exp(z_{S,j})}\right)
\end{equation}
where $C$ is the number of monitored websites.

\textbf{Resource-Privileged Distillation Loss.} To mitigate shortcut learning under ERM, we transfer the teacher's soft knowledge via KL divergence with temperature $\tau$:
\begin{equation}
    \mathcal{L}_{kd} = \tau^2 \cdot \mathcal{D}_{KL}\left( \sigma\left(\frac{z_T}{\tau}\right) \parallel \sigma\left(\frac{z_S}{\tau}\right) \right)
\end{equation}
where $\sigma(\cdot)$ is the Softmax function. The temperature $\tau$ flattens the posterior distribution, amplifying inter-class similarity signals beyond the target class. Minimizing this loss guides the student to internalize the inter-class relationships encoded by the resource modality, acting as a form of \textit{semantic regularization}.

\textbf{Joint Objective.} The student's total training objective is a weighted combination:
\begin{equation}
    \mathcal{L}_{total} = (1 - \alpha) \mathcal{L}_{cls} + \alpha \mathcal{L}_{kd}
\end{equation}
$\alpha \in [0, 1]$ controls the trade-off between the classification objective and the privileged distillation objective. At the boundary $\alpha = 0$, $\mathcal{L}_{total}$ reduces to $\mathcal{L}_{cls}$, and the student degenerates into a standard traffic-only classifier trained under ordinary ERM. In practice, the optimal $\alpha$ is primarily determined by the student backbone and exhibits relatively low sensitivity to the specific training and testing datasets; we analyze its effect in the ablation study (\S\ref{sec:appendix_alpha}).

Mechanistically, $\mathcal{L}_{cls}$ preserves the student's ability to discriminate ground-truth website labels, while $\mathcal{L}_{kd}$ answers: ``which websites share similar resource loading structures?'' Together, they prevent the student from relying solely on one-hot supervision, effectively suppressing overfitting to transient, environment-specific traffic patterns.

\subsection{Online Inference and Deployment}

After distillation, ResAware retains only the student model $S$ for deployment. All training-specific components, including the resource parser, the teacher model, and the distillation objective, are removed before deployment.

At inference time, the deployed model takes a single encrypted traffic trace as input, without expanding the online attack surface. All additional computation introduced by ResAware is confined to the offline training phase. As a result, the deployed model has the same inference latency and memory footprint as the underlying baseline, making ResAware a plug-and-play enhancement with zero additional online overhead.

\section{Evaluation}

This section evaluates the effectiveness, generality, and underlying mechanisms of ResAware under diverse cross-environment WF settings.

\subsection{Experimental Setup}

\noindent \textbf{Datasets and Evaluation Protocols.}
Since existing public WF datasets lack application-layer resource events synchronized with traffic traces, we collect a large-scale evaluation dataset of paired traffic-resource samples. Each sample is represented as $(x, x^*, y)$, where $x$ is the encrypted traffic trace, $x^*$ is the privileged resource sequence (accessible only at training time), and $y$ is the website label. Detailed resource sequence construction procedures are provided in Appendix~\ref{sec:a.1}.

Data collection spanned November 2025 to April 2026 across six geographically distributed vantage points (US, Japan, Singapore, South Africa, Australia, and Germany). The monitored set comprises 100 stable websites randomly sampled from the Tranco Top 100K~\cite{lepochat2019tranco}. The unmonitored set consists of 83,645 reachable background websites excluding the monitored set. In total, we collected over 160,000 page-load traces (collection pipelines and distribution statistics are detailed in Appendix~\ref{sec:a.2}). In the source domain, each monitored site comprises 150 traces used for model training; in each target-domain test set, each monitored site contributes 25--30 traces per snapshot. For open-world evaluation, the background split contains 1 trace per unmonitored site.

In all cross-environment experiments, target-domain samples are strictly excluded from model training, distillation hyperparameter selection, and threshold tuning. Distillation hyperparameters are tuned once on the source-domain validation set and fixed thereafter. All experiments are run five times with different random seeds; we report the mean performance. We design five evaluation scenarios to cover realistic deployment shifts:

\begin{table}[t]
\centering
\caption{Summary of backbone models, input feature representations, and the source-validated selected distillation weight $\alpha$ } 
\label{tab:backbone_models}
\resizebox{0.9\columnwidth}{!}{
\begin{tabular}{ccc}
\toprule
\textbf{Backbone} & \textbf{Input Features} & \textbf{$\alpha$} \\
\midrule
AWF~\cite{rimmer2018automated} & Packet direction sequence & 0.1 \\
DF~\cite{sirinam2018deepFingerprinting} & Packet direction sequence & 0.5 \\
RF~\cite{wang2014effectiveAttacks} & Traffic aggregated features & 0.5 \\
Var-CNN~\cite{bhat2019varcnn} & Packet direction sequence & 0.7 \\
Tik-Tok~\cite{rahman2020tiktok} & Packet direction and timestamp sequence & 0.5 \\
CountMamba~\cite{deng2025countmamba} & \makecell[c]{Packet direction, length, \\ and timestamp sequence} & 0.7 \\
\bottomrule
\end{tabular}
}
\end{table}

\begin{table*}[t]
\centering
\caption{Zero-shot closed-world F1-score with and without ResAware across four drift settings. $\Delta$ denotes the absolute gain in percentage points}
\label{tab:drift_comparison}
\small
\begin{tabular}{l|ccc|ccc|ccc|ccc}
\toprule
\multirow{2}{*}{Model} & \multicolumn{3}{c|}{Temporal Drift (Day 150)} & \multicolumn{3}{c|}{Spatial Drift (Avg.)} & \multicolumn{3}{c|}{Proxy Drift (Avg.)} & \multicolumn{3}{c}{Browser Drift (Avg.)} \\
\cmidrule(lr){2-4} \cmidrule(lr){5-7} \cmidrule(lr){8-10} \cmidrule(lr){11-13}
 & w/o & w/ & $\Delta$ & w/o & w/ & $\Delta$ & w/o & w/ & $\Delta$ & w/o & w/ & $\Delta$ \\
\midrule
AWF        & 33.25 & 32.25 & -1.00          & 49.23 & 48.76 & -0.47          & 17.53 & 18.03 & \textbf{+0.50} & 5.91  & 6.06  & \textbf{+0.15} \\
CountMamba & 28.94 & 29.16 & \textbf{+0.22} & 72.91 & 76.03 & \textbf{+3.12} & 61.21 & 62.50 & \textbf{+1.29} & 7.11  & 9.50  & \textbf{+2.39} \\
RF         & 36.64 & 38.27 & \textbf{+1.63} & 76.11 & 78.61 & \textbf{+2.50} & 62.86 & 66.74 & \textbf{+3.88} & 18.15 & 22.83 & \textbf{+4.68} \\
Tik-Tok     & 54.64 & 57.67 & \textbf{+3.03} & 82.85 & 85.10 & \textbf{+2.25} & 44.52 & 44.88 & \textbf{+0.36} & 4.79  & 6.05  & \textbf{+1.26} \\
DF         & 61.39 & 65.79 & \textbf{+4.40} & 84.71 & 86.64 & \textbf{+1.93} & 48.32 & 47.28 & -1.04          & 4.07  & 6.66  & \textbf{+2.59} \\
Var-CNN    & 72.77 & 81.49 & \textbf{+8.72} & 82.66 & 86.96 & \textbf{+4.30} & 38.14 & 47.10 & \textbf{+8.96} & 17.24 & 21.45 & \textbf{+4.21} \\
\bottomrule
\end{tabular}
\end{table*}

\begin{itemize}
    \item \textit{Temporal Drift.} Models are trained on the source domain and tested on temporal snapshots collected at $\sim$30-day intervals to evaluate cross-time generalization.
    \item \textit{Spatial Drift.} The test set consists of samples from five geographic locations within the same time window, assessing generalization across diverse network paths and CDN deployments.
    \item \textit{Obfuscation proxy Drift.} The test set covers six obfuscation proxy protocols (Shadowsocks~\cite{shadowsocks2016}, Trojan\cite{trojan_gfw_2023}, VLESS-XTLS-Vision, VMess-WS-TLS, VMess-TLS and VMess\cite{project_x_2020}) to evaluate resilience against transport-layer obfuscation.
    \item \textit{Browser Drift:} By designating Chrome as the source domain for training and utilizing Edge and Firefox as target domains, we evaluate the robustness against variations in rendering engines and connection management.
    \item \textit{Open-World Temporal Drift.} Using temporal snapshots, we mix 100 monitored classes (8 samples each) with 80,000 unmonitored classes(1 samples each) at a 1:100 ratio. This scenario evaluates detection capability under strict false-positive constraints ($TPR@1\%FPR$).
\end{itemize}

\noindent \textbf{Backbone Models.}
To cover diverse input representations and modeling paradigms, we select six representative WF architectures as student backbones. Table~\ref{tab:backbone_models} lists the input features and validated $\alpha$ values for each model; the optimal $\alpha$ is coupled to the model architecture, a relationship we analyze in \S\ref{sec:appendix_alpha}. For each baseline, we compare its native version against its ResAware-distilled counterpart.

\noindent \textbf{Teacher Model and Reporting Role.}
Unless otherwise specified, ``Teacher'' refers exclusively to the resource-only teacher defined in \S\ref{sec:training_teacher_model}. This Transformer-based model is trained on the same source-domain monitored websites as the student models, using only resource sequences. Because it relies on a privileged modality unavailable to the online attacker, its predictions are reported only as an oracle-style diagnostic reference for resource-side stability; they are neither a deployable baseline nor a formal upper bound for traffic-side student models.

\noindent \textbf{Training Protocol and Hyperparameter Fairness.}
To ensure that performance gains are attributable to ResAware rather than additional hyperparameter tuning, we follow the architectures, optimizers, learning rate schedules, batch sizes, and training epochs from the original papers or official implementations for each backbone. For the native and ResAware-distilled versions of the same baseline, all training configurations are kept identical except for the distillation mechanism itself. ResAware-specific hyperparameters (temperature and distillation weight $\alpha$) are tuned once on the source-domain validation set and fixed for all subsequent experiments; they are never re-tuned for target environments.

\noindent \textbf{Metrics.}
For closed-world tasks, we adopt the F1-score as the primary evaluation metric, supplemented by Precision and Recall. For open-world tasks, given the attacker's sensitivity to false alarms, we prioritize True Positive Rate at a fixed False Positive Rate ($TPR@1\%FPR$) as the primary metric.

\noindent \textbf{Implementation Details.}
We implement ResAware in Python 3.12 with PyTorch 2.10.0. All training, distillation, and inference experiments run on a single workstation (Ubuntu 24.04 LTS) equipped with dual Intel Xeon Platinum 8352S CPUs, 128 GB RAM, and an NVIDIA RTX 4090 GPU (24 GB VRAM). Unless otherwise noted, all backbone implementations use the same random seeds to ensure reproducibility.

\subsection{Zero-Shot Robustness under Cross-Environment Drift}
\label{sec:zero-shot}

We evaluate whether ResAware improves the zero-shot robustness of WF models under four cross-environment drift scenarios, where models are trained on the source domain with no access to target-domain samples.

\noindent \textbf{Overall Results.}
Table~\ref{tab:drift_comparison} summarizes the zero-shot closed-world F1-scores across six backbone models (see Appendix~\ref{sec:all_result} for per-environment breakdowns). ResAware yields positive gains in 21 of 24 backbone $\times$ drift combinations (87.5\%).

\begin{figure*}[t]
    \centering
    \includegraphics[width=0.92\textwidth]{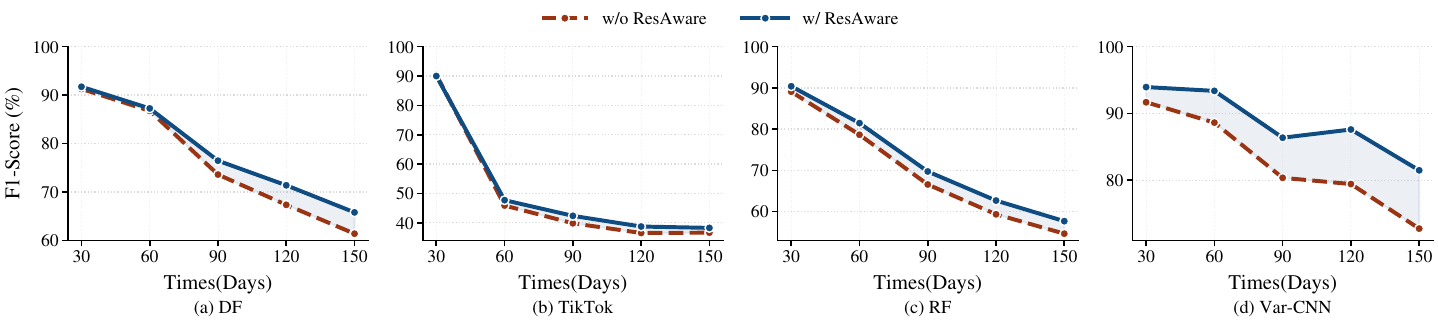}
    \caption{Closed-world F1-score (\%) with and without ResAware over five temporal test snapshots (Day 30--150) for WF backbones.}
    \label{fig:time_result}
\end{figure*}

\noindent \textbf{Temporal Drift.}
Under a 150-day drift, ResAware improves Var-CNN from 72.77\% to 81.49\% (+8.72\%), with additional gains of +4.40\% for DF and +3.03\% for Tik-Tok. The temporal decay curves in Figure~\ref{fig:time_result} show that the performance gap generally widens over longer intervals for the stronger sequential backbones, indicating that ResAware slows degradation under long-horizon drift rather than merely improving source-domain fit. Table~\ref{tab:temporal_prf} further reports per-snapshot Precision, Recall, and F1 for all six backbones; the ResAware student (Var-CNN backbone) remains close to the Teacher across all five snapshots, dropping only from 93.95\% at Day~30 to 81.49\% at Day~150, while the vanilla Var-CNN baseline falls to 72.77\%. Notably, the Teacher model maintains high accuracy after 150 days, confirming that page-level resource organization is substantially more stable over time than packet morphology. ResAware exploits this asymmetry to regularize the student's decision boundaries.

\begin{table*}[t]
\centering
\caption{Precision, Recall, and F1-score (\%) under temporal drift for all six backbones and the resource-only teacher across five test snapshots (Day 30--150).}
\label{tab:temporal_prf}
\resizebox{0.95\textwidth}{!}{
\setlength{\tabcolsep}{5pt}
\begin{tabular}{l|ccc|ccc|ccc|ccc|ccc}
\toprule
 & \multicolumn{3}{c|}{\textbf{Day 30}} & \multicolumn{3}{c|}{\textbf{Day 60}} & \multicolumn{3}{c|}{\textbf{Day 90}} & \multicolumn{3}{c|}{\textbf{Day 120}} & \multicolumn{3}{c}{\textbf{Day 150}} \\
\cmidrule(lr){2-4} \cmidrule(lr){5-7} \cmidrule(lr){8-10} \cmidrule(lr){11-13} \cmidrule(lr){14-16}
 & \textbf{P} & \textbf{R} & \textbf{F1} & \textbf{P} & \textbf{R} & \textbf{F1} & \textbf{P} & \textbf{R} & \textbf{F1} & \textbf{P} & \textbf{R} & \textbf{F1} & \textbf{P} & \textbf{R} & \textbf{F1} \\ \midrule
\textit{Teacher} & \textit{97.35} & \textit{96.77} & \textit{96.49} & \textit{95.55} & \textit{94.77} & \textit{94.44} & \textit{89.74} & \textit{90.93} & \textit{89.27} & \textit{89.61} & \textit{89.57} & \textit{87.61} & \textit{90.49} & \textit{98.80} & \textit{88.97} \\ \midrule
AWF & 55.89 & 54.05 & 51.23 & 52.78 & 51.64 & 48.18 & 43.72 & 44.30 & 40.14 & 41.86 & 42.94 & 38.20 & 35.77 & 37.95 & 33.25 \\
CountMamba & 88.39 & 89.31 & 87.79 & 44.73 & 36.71 & 35.26 & 38.19 & 33.39 & 31.79 & 36.35 & 31.93 & 29.79 & 34.73 & 31.36 & 28.94 \\
Tik-Tok & 90.30 & 90.15 & 89.07 & 80.45 & 80.97 & 78.63 & 69.46 & 69.51 & 66.57 & 63.24 & 62.38 & 59.34 & 58.40 & 57.83 & 54.64 \\
RF & 91.25 & 91.52 & 90.46 & 61.96 & 47.12 & 45.88 & 52.24 & 41.82 & 39.84 & 49.83 & 38.84 & 36.53 & 47.84 & 39.47 & 36.64 \\
DF & 91.99 & 92.08 & 91.28 & 88.17 & 88.33 & 86.75 & 74.82 & 76.90 & 73.60 & 68.60 & 71.65 & 67.35 & 62.08 & 66.19 & 61.39 \\
Var-CNN & 92.90 & 92.52 & 91.68 & 90.30 & 89.56 & 88.63 & 83.57 & 82.03 & 80.37 & 82.33 & 81.62 & 79.44 & 74.51 & 76.22 & 72.77 \\ \midrule
\textbf{ResAware} & \textbf{94.11} & \textbf{94.66} & \textbf{93.95} & \textbf{94.22} & \textbf{94.00} & \textbf{93.38} & \textbf{87.93} & \textbf{87.81} & \textbf{86.37} & \textbf{89.46} & \textbf{89.19} & \textbf{87.60} & \textbf{82.57} & \textbf{84.28} & \textbf{81.49} \\ \bottomrule
\end{tabular}
}
\end{table*}

\noindent \textbf{Spatial Drift.}
Across the five international vantage points, ResAware improves five of the six backbones on average. The largest gains appear for Var-CNN (+4.30\%, 82.66\%$\to$86.96\%), followed by CountMamba (+3.12\%) and RF (+2.50\%), while DF and Tik-Tok also improve by +1.93\% and +2.25\%, respectively. AWF is the only exception, showing a slight average drop (-0.47\%), which indicates that spatial drift is generally mild enough for resource supervision to help, but low-capacity students may still fail to absorb the transferred topology consistently.

\noindent \textbf{Obfuscated Proxy Drift.}
Obfuscation Proxy protocols heavily distort packet-level morphology while leaving page resource structure largely intact. The largest improvement is observed for Var-CNN, whose F1 score increases from 38.14\% to 47.10\% (+8.96 \%), followed by RF (+3.88 \%) and CountMamba (+1.29 \%). AWF and Tik-Tok obtain only marginal gains (+0.50 and +0.36 \%), while DF is the only backbone with a slight degradation (-1.04 \%). This exception indicates that, under severe proxy-induced deformation, cross-modal distillation is not uniformly beneficial across architectures; its effectiveness still depends on the student's ability to align traffic representations with the resource-level topology transferred by the teacher.

\noindent \textbf{Cross-Browser Drift.}
Browser drift is the most challenging of the four scenarios: under the vanilla setting, four of the six backbones remain below 10\% average F1-score, with only RF (18.15\%) and Var-CNN (17.24\%) retaining limited discriminative power. ResAware still yields consistent gains for all six backbones, led by RF (+4.68\%), Var-CNN (+4.21\%), and DF (+2.59\%), although the absolute performance remains far below that under temporal and spatial drift. This result suggests that browser switching perturbs how application resources are rendered, scheduled, and multiplexed into observable traffic, making cross-modal topology transfer substantially harder than in the other drift settings.

\noindent \textbf{Takeaways.}
The zero-shot results support two conclusions: (1) Resource supervision during training improves most traffic-only WF models, with the clearest and most stable benefits under long-term temporal drift. (2) ResAware is not an unconditional enhancer: its effectiveness depends on whether resource sequences remain predictive of observable traffic morphology and whether the student has sufficient capacity to absorb the teacher's inter-class topology; when browser execution or proxy encapsulation disrupts this correspondence, or when the student cannot accommodate the distillation constraint, gains may become limited or turn into negative transfer (\S\ref{sec:applicability}).

\subsection{Open-World Detection under Temporal Drift}

\begin{table}[t]
\centering
\caption{Open-world temporal drift results ($TPR@1\%FPR$) for DF, Tik-Tok, and Var-CNN with and without ResAware across five temporal snapshots (100 monitored sites vs.\ 80K unmonitored, 1:100 ratio). $\Delta$ denotes the absolute gain in percentage points.}
\label{tab:open_world_temporal_modified}
\resizebox{0.9\columnwidth}{!}{
\begin{tabular}{l l ccccc}
\toprule
\textbf{Model} & \textbf{Type} & \textbf{30} & \textbf{60} & \textbf{90} & \textbf{120} & \textbf{150} \\
\midrule
\multirow{3}{*}{DF}      & w/o   & 50.68 & 39.52 & 28.45 & 24.87 & 20.40 \\
                         & w/    & 54.23 & 40.70 & 28.63 & 26.00 & 21.02 \\
                         & \textbf{$\Delta$} & \textbf{+3.55} & \textbf{+1.18} & \textbf{+0.18} & \textbf{+1.13} & \textbf{+0.62} \\
\midrule
\multirow{3}{*}{Tik-Tok}  & w/o   & 27.50 & 8.60  & 6.73  & 5.50  & 4.52  \\
                         & w/    & 50.15 & 22.63 & 16.43 & 12.93 & 10.17 \\
                         & \textbf{$\Delta$} & \textbf{+22.65} & \textbf{+14.03} & \textbf{+9.70} & \textbf{+7.43} & \textbf{+5.65} \\
\midrule
\multirow{3}{*}{Var-CNN} & w/o   & 48.75 & 35.70 & 27.57 & 24.92 & 22.40 \\
                         & w/    & 55.07 & 41.05 & 30.43 & 28.85 & 27.20 \\
                         & \textbf{$\Delta$} & \textbf{+6.32} & \textbf{+5.35} & \textbf{+2.86} & \textbf{+3.93} & \textbf{+4.80} \\
\bottomrule
\end{tabular}
}
\end{table}

Closed-world performance alone is insufficient to assess the practical threat of WF attacks; we therefore evaluate ResAware under the open-world temporal drift setting. We focus on the three strongest backbones in this regime, namely DF, Tik-Tok, and Var-CNN, and report TPR at a stringent operating point of 1\% FPR under the 1:100 monitored-to-unmonitored imbalance.

Table~\ref{tab:open_world_temporal_modified} shows that ResAware improves TPR@FPR=0.01 for all three backbones across the full 150-day window. The gains are most pronounced for Tik-Tok, where TPR rises by +22.65\%, +14.03\%, and +9.70\% over the first three snapshots, and remains +5.65\% higher even at Day~150. Var-CNN exhibits consistently positive improvements (+2.86\% to +6.32\%), while DF remains comparatively robust and still benefits from modest gains (+0.18\% to +3.55\%). Even when the baseline detector degrades substantially under long-term drift, resource supervision preserves meaningful monitored-site detection capability at a strict false-positive budget.

\noindent \textbf{Takeaway.}
The closed-world robustness gains from ResAware carry over to the more operationally relevant open-world setting: training-time resource supervision improves low-FPR monitored-site detection under temporal aging and severe class imbalance, with the largest benefits appearing in backbones whose traffic-side decision boundaries are otherwise most vulnerable to long-term drift.

\subsection{Target-Domain Data Efficiency: Few-Shot and Zero-Label Adaptation}
\label{sec:target}
\begin{figure}[t]
    \centering
    \includegraphics[width=0.95\linewidth]{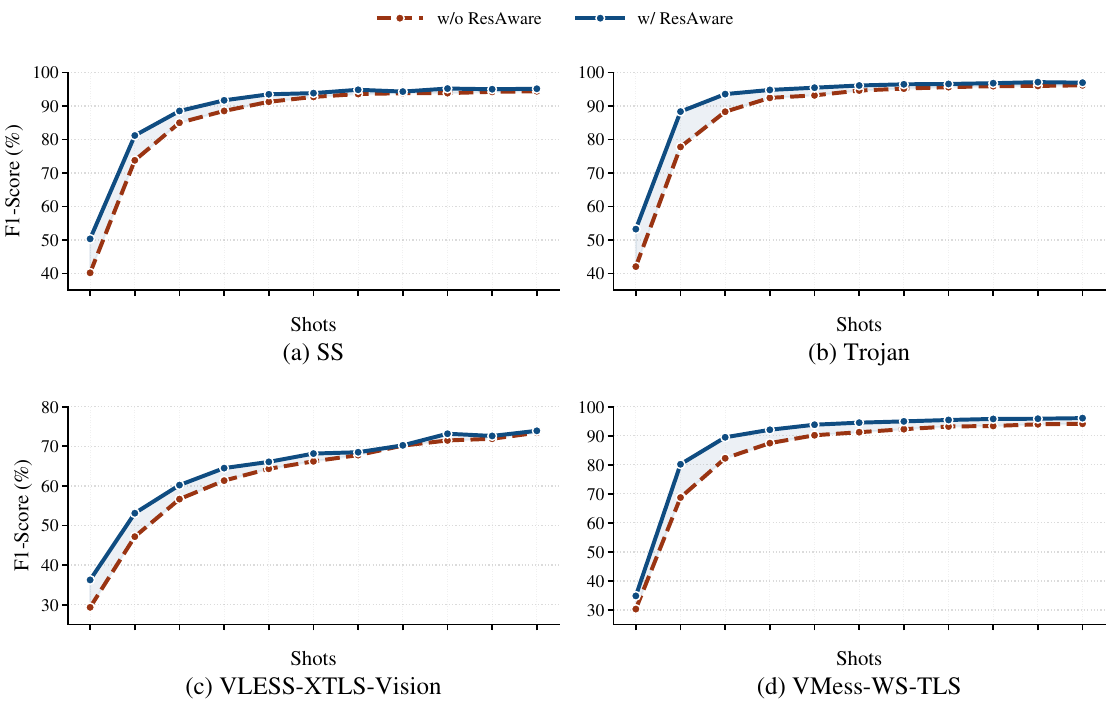}
    \caption{Few-shot adaptation F1-score (\%) of Var-CNN with and without ResAware under Shadowsocks, Trojan, VLESS-XTLS-Vision and VMess-WS-TLS proxy drifts.}
    \label{fig:few_result}
\end{figure}

In practice, WF attackers may occasionally obtain limited target-domain information after initial deployment. Such information takes two forms: a stronger but more costly variant in which the attacker acquires a small number of labeled target traces via controlled probing or repeated visits~\cite{sirinam2019triplet, fewshot2021chen}, and a weaker but more scalable variant in which the attacker observes unlabeled target traffic for Test-Time Adaptation (TTA~\cite{wang2021tent, crossdomain2020li}). Using Var-CNN as the backbone, we evaluate whether ResAware reduces the target-domain data required to recover performance under distribution shift.

\noindent \textbf{Few-Shot Adaptation with Labeled Target Samples.}
Following the standard few-shot evaluation protocol adopted in prior WF and domain adaptation works~\cite{sirinam2019triplet, fewshot2021chen}, we freeze the backbone and update only the linear classifier with $K$ labeled target-domain traces. Figure~\ref{fig:few_result} shows that ResAware provides the largest benefits in the low-shot regime. Under the Trojan proxy, ResAware achieves 88.33\% with just 1 shot, whereas vanilla Var-CNN reaches only 77.78\%. Under the more disruptive VMess-WS-TLS setting, ResAware with 5 shots matches the performance of vanilla Var-CNN with 10 shots (94.50\% vs.\ 94.11\%), effectively halving the label requirement.

\noindent \textbf{Unlabeled Target-Domain Adaptation.}
We further assess ResAware’s compatibility with Proteus~\cite{deng2026proteus}, a state-of-the-art unlabeled adaptation framework, under six obfuscation proxy drift settings (Table~\ref{tab:unlabeled_adaptation}). ResAware proves to be highly complementary to adaptation techniques. Proteus elevates vanilla Var-CNN accuracy from 38.79\% to 54.89\%, while ResAware + Proteus achieves a superior 69.14\%. This complementarity reflects their distinct operational stages: ResAware focuses on environment-agnostic representation learning in the source domain, whereas Proteus facilitates target-domain calibration. Consequently, ResAware serves as a stronger feature initializer rather than a substitute for test-time adaptation.

\noindent \textbf{Takeaways.}
ResAware improves the efficiency of both labeled and unlabeled target-domain adaptation: it reduces the labeled sample requirement for few-shot adaptation and provides a more robust feature initialization for unlabeled adaptation.

\begin{table}[t]
  \centering
  \caption{Closed-world F1-score (\%) under Obfuscation proxy drift for Var-CNN across six obfuscation protocols, under four configurations of ResAware and Proteus~\cite{deng2026proteus}. ResAware operates at training time (source-side); Proteus operates at inference time (target-side); their combination consistently outperforms either component alone.}
  \label{tab:unlabeled_adaptation}
  \resizebox{0.9\columnwidth}{!}{
  \begin{tabular}{l c c c c}
    \toprule
    \textbf{ResAware} & w/o & w/o & w/ & w/ \\
    \textbf{Proteus}  & w/o & w/ & w/o & w/ \\
    \midrule
    Shadowsocks         & 40.87\% & 56.43\% & 49.83\% & 74.70\% \\
    Trojan              & 43.07\% & 60.72\% & 52.96\% & 82.32\% \\
    VLESS-XTLS-Vision   & 29.75\% & 34.47\% & 34.98\% & 39.86\% \\
    VMess               & 46.28\% & 64.87\% & 55.58\% & 85.04\% \\
    VMess-TLS           & 41.88\% & 62.96\% & 51.33\% & 75.25\% \\
    VMess-WS-TLS        & 30.90\% & 49.87\% & 34.35\% & 57.66\% \\
    \midrule
    \textbf{AVG}        & \textbf{38.79\%} & \textbf{54.89\%} & \textbf{46.51\%} & \textbf{69.14\%} \\
    \bottomrule
  \end{tabular}
  }
\end{table}

\subsection{Ablation Analysis: What Makes ResAware Work?}
\label{sec:alpha_sensitivity}

Prior experiments establish that ResAware consistently improves the robustness of traffic-only WF models across diverse distribution shifts. We now investigate the sources of these gains through two ablation studies. First, we verify whether the improvements stem from correctly aligned privileged resource supervision or merely from the regularization effect of soft-label distillation. Second, we ablate individual resource channels—size, category, and order—to quantify each channel's contribution. The sensitivity analysis of the distillation weight $\alpha$ is provided in Appendix~\ref{sec:appendix_alpha}.

\begin{table}[t]
\centering
\caption{Ablation study verifying the necessity of correctly aligned privileged supervision under 150-day temporal drift (F1-score (\%)). Three conditions are compared: ResAware with a resource teacher, KD with a traffic teacher, and KD with class-shuffled resource soft labels.}
\label{tab:necessity-supervision}
\small
\setlength{\tabcolsep}{0pt}
\begin{tabular*}{\columnwidth}{@{\extracolsep{\fill}}lcccc}
\toprule
\textbf{Model} &
\makecell[c]{\textbf{w/ Resource} \\ \textbf{KD (Ours)}} &
\textbf{Baseline} &
\makecell[c]{\textbf{w/ Traffic} \\ \textbf{KD}} &
\makecell[c]{\textbf{w/ Class-Shuffled} \\ \textbf{Resource KD}} \\ \midrule

\textit{Teacher}    & \textit{88.97\%} & \textit{-}       & \textit{77.15\%} & \textit{-}     \\ \midrule
AWF        & 32.25\% & 33.25\% & 31.02\% & 29.44\% \\
DF         & 65.79\% & 61.39\% & 55.31\% & 62.58\% \\
RF         & 38.27\% & 36.64\% & 30.98\% & 37.71\% \\
Tik-Tok     & 57.67\% & 54.64\% & 40.61\% & 54.52\% \\
Var-CNN    & 81.49\% & 72.77\% & 51.48\% & 74.92\% \\
CountMamba & 29.16\% & 28.94\% & 24.36\% & 28.11\% \\ \bottomrule
\end{tabular*}
\end{table}

\noindent \textbf{Privileged Resource KD vs.\ Generic KD.}
We design two control conditions to isolate the source of gains: \textit{Traffic KD} replaces the teacher's input with traffic burst features (retaining the distillation pipeline but removing the resource modality), and \textit{Class-Shuffled Resource KD} preserves the numerical distribution of the resource soft labels but randomly permutes the class assignments (to test whether gains arise solely from soft-label regularization).

As shown in Table~\ref{tab:necessity-supervision}, both control conditions perform substantially worse than ResAware and generally fall below the baseline. Traffic KD not only fails to improve robustness but exacerbates degradation, indicating that a same-modality teacher reinforces the student's reliance on source-domain-specific spurious correlations. Class-Shuffled Resource KD performs at or below the baseline, ruling out soft-label regularization as the primary driver of gains. These results confirm that the student inherits the correct inter-class topology from the resource teacher, not merely the numerical smoothing of its soft labels.

\begin{figure}[t]
    \centering
    \includegraphics[width=\linewidth]{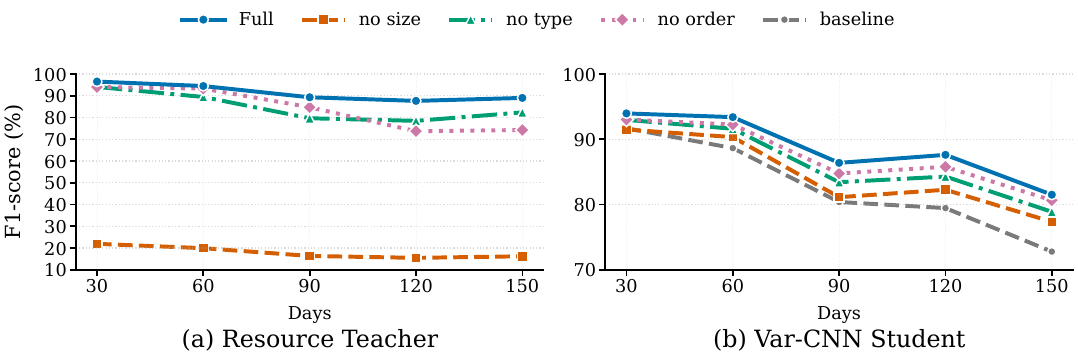}
    \caption{Per-channel ablation F1-score (\%) for the resource-only teacher and ResAware Var-CNN student across temporal drift (Days 30–150).}
    \Description{}
    \label{fig:resource_feature}
\end{figure}

\noindent \textbf{Contribution of Resource Channels.}
We evaluate the teacher and Var-CNN student under temporal drift by selectively ablating resource size (\textit{no size}), type (\textit{no type}), or request order (\textit{no order}) to quantify each channel's contribution.

As shown in Figure~\ref{fig:resource_feature}, resource size is the strongest discriminative signal: removing it drops the teacher's 150-day F1-score from 88.97\% to 16.16\%, while the student drops from 81.49\% to 77.34\%. Resource type and order are nonetheless non-redundant: ablating type incurs a 6.65\% F1-score drop for the teacher under 150-day drift, and shuffling order causes a 14.76\% drop. Even when distilling from only partial resource channels (size and type), the student's cross-environment robustness remains above the traffic-only baseline. All resource channels thus encode transferable, stable structural information.

\noindent \textbf{Sensitivity of Distillation Weight $\alpha$.}
The weight $\alpha \in [0,1]$ balances hard-label classification and resource-topology supervision in $\mathcal{L}_{total}$. Across backbone and capacity sweeps (Figure~\ref{fig:alpha_ab} and Table~\ref{tab:capacity_scaling}), the near-optimal range is primarily capacity-dependent and stable across training and testing datasets; we therefore tune $\alpha$ once per backbone on the source-domain validation set and fix it for all target environments, with the full analysis deferred to Appendix~\ref{sec:appendix_alpha}.

\noindent \textbf{Takeaways.}
The ablation studies confirm three points: (1) Gains stem from correctly aligned privileged resource knowledge, not from the distillation mechanism or soft-label regularization alone; (2) Resource size is the strongest single channel, while type and order provide complementary structural constraints; (3) $\alpha$ is a capacity-matching parameter for each student backbone, calibrated once on the source-domain validation set (full analysis in Appendix~\ref{sec:appendix_alpha}).

\subsection{Applicability Analysis: When Does ResAware Fail?}
\label{sec:applicability}

The preceding experiments show that ResAware is not an unconditional plug-in enhancer. Its effectiveness relies on two conditions. First, resource sequences must continue to encode stable website identity across the source and target domains. Second, the student must have sufficient capacity to compress the resource teacher's inter-class topology into a \textit{traffic-only} representation. When either condition is weakened, the distillation term may provide only limited benefit; when both are violated, it can induce negative transfer.

\noindent \textbf{Failure from Broken Traffic-Resource Correspondence.}
The first failure mode arises when the correspondence between resource structure and observable traffic morphology is substantially disrupted. ResAware is most suitable for temporal and spatial drift, where the perturbation mainly affects the network-observation layer while the resource set, category sequence, and size distribution remain comparatively stable. In contrast, browser drift changes page-load scheduling, connection reuse, preloading behavior, and script execution order. Obfuscation proxy drift can also systematically rewrite the projection from resource events to traffic packet sequences through tunnel multiplexing, fragmentation, outer TLS encapsulation, or WebSocket framing. As a result, ResAware still yields relative gains under browser drift, but the absolute macro-F1 remains low. Obfuscation proxy drift also exhibits clear model dependence: Var-CNN and RF benefit, whereas DF shows slight negative transfer. These results indicate that once the target shift enters the browser execution layer or the protocol encapsulation layer, the teacher's resource-side soft labels may become a mismatched constraint rather than a stable prior.

\noindent \textbf{Failure from Insufficient Student Capacity.}
The second failure mode comes from limited student capacity. ResAware does not expose resource features to the student at inference time; instead, it asks a traffic-only student to fit both hard-label decision boundaries and the resource teacher's soft topology. The Var-CNN width-scaling experiment in Appendix~\ref{sec:appendix_alpha} shows that smaller students have narrower best $\alpha$ ranges and lower gain ceilings. The full-width Var-CNN maintains 80.25\%--82.22\% macro-F1 within the best range of $\alpha=0.1$--$0.7$, reaching a maximum gain of 9.45 percentage points. In contrast, the $0.125\times$ Var-CNN has a best range of only $\alpha=0.1$--$0.3$, with a maximum gain of 4.60 percentage points. Thus, low-capacity students are not unable to benefit from resource supervision; they simply absorb a weaker teacher constraint. An overly large $\alpha$ turns the KD term from structural regularization into an optimization burden.

\noindent \textbf{Deployment Guidelines.}
Based on the above analysis, we derive three practical guidelines. First, when drift mainly occurs below the resource layer, such as temporal aging, geographic relocation, CDN routing changes, or link-state variation, ResAware is a suitable default training enhancement. When the drift involves browser execution or complex proxy encapsulation, it should be validated per scenario, with the $\alpha=0$ traffic-only baseline retained as a fallback. Second, $\alpha$ should be matched to student capacity and inductive bias: DF, Tik-Tok, and RF benefit most from moderate weights; Var-CNN and CountMamba benefit from medium-to-high weights; AWF should use smaller weights and be checked for negative transfer. Third, deployment should retain only the traffic-only student, with no resource parser or teacher model in the inference pipeline. If a small amount of target-domain traffic is available, ResAware is best used as a stronger source-domain initialization that can be combined with few-shot or unlabeled adaptation.

\subsection{Mechanism Analysis: What Does the Student Inherit?}
\label{sec:mechanism_analysis}

\begin{figure}[t]
    \centering
    \includegraphics[width=\linewidth]{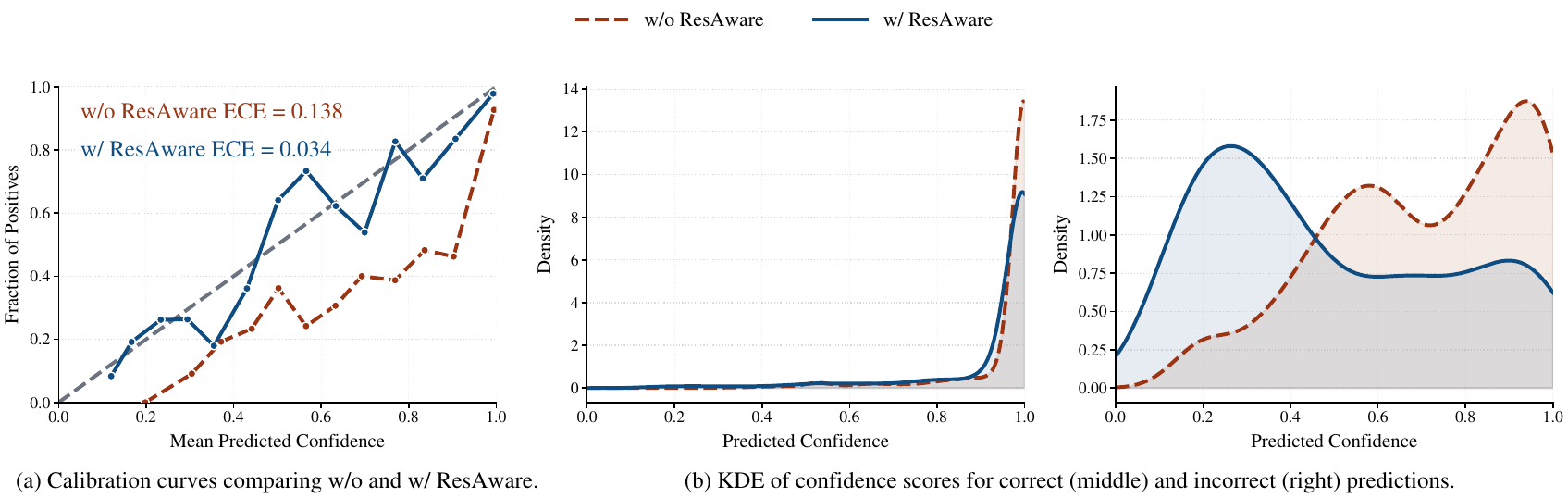}
    \caption{Calibration and confidence distributions of Var-CNN at Day~150 with and without ResAware.
    The left panel shows the reliability diagram, while the middle and right panels show the confidence KDEs for correct and incorrect predictions, respectively.}
    \label{fig:ece}
\end{figure}

This section investigates a core mechanistic question: what information does the resource teacher transfer to the traffic-only student through heterogeneous distillation? We find that the student inherits the resource-induced inter-class topology and acquires more stable decision boundaries.

\noindent \textbf{ResAware Improves Model Robustness and Calibration.}
Figure~\ref{fig:ece} characterizes the effect of ResAware on Var-CNN's output distribution at Day~150. The reliability diagram~\cite{guo2017calibration} shows that the baseline exhibits severe overconfidence (ECE\,=\,0.138), whereas ResAware reduces ECE to 0.034—a nearly fourfold improvement. The confidence KDE reveals a complementary pattern: for correct predictions, ResAware produces a sharper, more concentrated peak near 1.0, indicating higher decisiveness; for incorrect predictions, the baseline clusters errors near high-confidence regions, whereas ResAware shifts the error mass toward lower confidence.

These results show that ResAware does not uniformly suppress confidence; instead, it achieves structural calibration—being more confident when correct and more conservative when wrong. The teacher thus imprints resource-side structural invariants onto the student, guiding it away from overfitting to transient traffic noise.

\begin{table}[t]
\centering
\caption{Inter-class topology alignment between Var-CNN with and without ResAware and the resource teacher over 150 days. KL divergence ($\downarrow$) measures the distributional distance between student and teacher soft outputs; Spearman $\rho$ ($\uparrow$) measures the rank correlation of per-class similarity orderings.}
\label{tab:topology}
\resizebox{0.8\columnwidth}{!}{
\begin{tabular}{ccccc}
\toprule
\multirow{2}{*}{\textbf{Days}}  & \multicolumn{2}{c}{\textbf{KL to Teacher ($\downarrow$)}} & \multicolumn{2}{c}{\textbf{Rel. Spearman $\rho$ ($\uparrow$)}} \\
\cmidrule(lr){2-3} \cmidrule(lr){4-5}
& \textbf{w/o Res.} & \textbf{w/ Res.} & \textbf{w/o Res.} & \textbf{w/ Res.} \\
\midrule
30  & 2.5516 & 0.3838 & 0.0373 & 0.0806 \\
60  & 2.4311 & 0.3522 & 0.0273 & 0.0775 \\
90  & 2.2566 & 0.3261 & 0.0324 & 0.0860 \\
120 & 2.1713 & 0.3140 & 0.0297 & 0.0962 \\
150 & 2.1133 & 0.3029 & 0.0287 & 0.1022 \\
\bottomrule
\end{tabular}
}
\end{table}

\noindent \textbf{The Student Aligns with the Teacher's Soft Topology.}
To quantify how much of the teacher's inter-class topology the student inherits, we track two metrics over the 150-day drift window. \textit{KL divergence} measures the distributional distance between the student's and teacher's soft output distributions: a lower value indicates that the student assigns similar per-class probability mass as the teacher. \textit{Spearman $\rho$} measures the rank correlation between the student's and teacher's per-class similarity orderings: a higher value indicates that the student preserves the teacher's relative inter-class structure~\cite{knowledge2022huang}.

As shown in Table~\ref{tab:topology}, Var-CNN without ResAware remains far from the teacher throughout the window, with KL divergence in the range of 2.1--2.6 and Spearman $\rho$ near zero (0.027--0.037). In contrast, Var-CNN with ResAware maintains substantially closer alignment: KL divergence stays within 0.30--0.38, and Spearman $\rho$ increases steadily from 0.081 at Day~30 to 0.102 at Day~150. These results confirm that the student inherits the resource teacher's inter-class soft topology rather than simply mimicking hard-label predictions.

\noindent \textbf{Takeaway.}
ResAware converts resource-side structural priors from the training phase into stable relational supervision signals. At inference, the student relies exclusively on encrypted traffic, yet its decision boundaries are regularized by the resource-induced topology—enabling both higher F1-score and better calibration under long-term drift.

\section{Limitations and Future Work}

ResAware assumes that resource structure retains adequate stability in the target environment. For highly personalized pages, sites under frequent A/B testing, heavily ad-injected platforms, or dynamically generated templates, resource sizes, categories, and loading sequences may fluctuate substantially, weakening the structural priors provided by the teacher. Future work could explore more abstract resource representations—such as dependency graphs, initiator graphs, or rendering-stage topologies—to reduce sensitivity to specific object sizes.

We do not evaluate strong anonymity networks such as Tor. Tor's fixed-size cells, multiplexing, and congestion control further obscure the correspondence between application-layer resource sizes and observable packet sequences. Extending ResAware to such networks would likely require a resource teacher that de-emphasizes object sizes in favor of sequential or graph-based representations.

Finally, ResAware does not uniformly benefit all backbone--drift combinations. As shown in \S\ref{sec:zero-shot} and \S\ref{sec:applicability}, severe browser or obfuscated proxy drift can break the traffic-resource correspondence, while low-capacity students may fail to absorb the teacher's inter-class topology, leading to diminished gains or negative transfer. Future work should develop capacity-aware weighting and drift-aware criteria for falling back to traffic-only training when resource supervision becomes unreliable.
\section{Related Work}

\noindent \textbf{Website Fingerprinting Attacks.}
Deep learning-based WF models, including DF~\cite{sirinam2018deepFingerprinting}, Var-CNN~\cite{bhat2019varcnn}, Tik-Tok~\cite{rahman2020tiktok}, and CountMamba~\cite{deng2025countmamba}, achieve high accuracy in closed-world IID settings by learning low-level packet statistics. However, packet-level signals are jointly shaped by website content structure \emph{and} transport-layer dynamics: TCP congestion control, HTTP/2 multiplexing, CDN routing, and browser scheduling all modulate observable traffic independently of the page being loaded. Consequently, even modest changes in network path, obfuscated proxy protocol, or browser engine suffice to invalidate learned patterns~\cite{cherubin2022online, shusterman2026conceptDrift, li2025crossEnvironmental, deng2025singlePerspective, shadbeh2026realityCheck}, reflecting a fundamental mismatch between the stability of application-layer content structure and the volatility of its encrypted traffic projection~\cite{critical2014juarez}.

\noindent \textbf{Improving Robustness within the Traffic-Only Threat Model.}
Existing efforts fall into two categories. The first improves traffic feature representations: Shen et al.~\cite{shen2023robustTrafficRepresentation} use feature attribution and contrastive regularization to suppress defense-induced perturbations; Bahramali et al.~\cite{bahramali2023realistic} simulate network-condition variation via trace augmentation; and Shen et al.~\cite{shen2025swallow} pursue transfer-robust training objectives. While each reduces sensitivity to a specific perturbation type, all remain constrained by working within the encrypted traffic domain, since the instability of these signals stems from sources outside the traffic itself, limiting the reach of any representation-level fix.

The second category introduces post-deployment target-domain observations. Few-shot adaptation methods fine-tune the classifier with a small number of labeled target traces~\cite{sirinam2019triplet, fewshot2021chen, zou2022brownianDistance}; test-time adaptation methods operate on unlabeled target traffic via entropy minimization or distribution alignment~\cite{zhang2023unsupervised, deng2026proteus}. Both improve accuracy under drift but require post-deployment data collection and do not eliminate the root cause: the adapted model still anchors its decision boundaries to volatile traffic-domain signals. ResAware is orthogonally complementary to these methods (\S\ref{sec:target}): by providing a more stable source-domain initialization, it amplifies the benefit of subsequent adaptation rather than replacing it.

\noindent \textbf{Resource-Aware Website Fingerprinting.}
A separate research thread exploits application-layer resource structure as a more stable website identifier. Li et al.~\cite{li2023robust} show that resource loading sequences exhibit substantially greater cross-environment stability than packet features. HOLMES \& WATSON~\cite{cheng2025holmesWatson} infers HTTP parallelism patterns directly from traffic as lightweight fingerprints; MRCGCN~\cite{gao2025mrcgcn} constructs multi-level resource dependency graphs; and STAR~\cite{cheng2025star} trains dual encoders to align traffic and resource representations for zero-shot cross-modal retrieval. These works confirm that resource-level signals offer a more stable encoding of website identity. However, their deployment assumptions differ: HOLMES infers structural signals from traffic alone and therefore needs no resource access at inference, but is bounded by what traffic can reveal about resource structure. MRCGCN and STAR directly incorporate resource graphs or embeddings that must be available at inference time, expanding the attacker's observational requirement beyond standard passive eavesdropping. ResAware takes a different position: resource information is used exclusively as a privileged training-time supervision signal and fully discarded before deployment, leaving the online model with the same footprint as a conventional traffic-only classifier.

\noindent \textbf{Learning Using Privileged Information and Knowledge Distillation.}
Vapnik and Izmailov~\cite{vapnik2015privileged,newlearning2009vapnik} formalize Learning Using Privileged Information (LUPI): auxiliary features available only at training time can substitute for larger datasets by providing richer concept supervision. Lopez-Paz et al.~\cite{lopezpaz2016unifying} establish a formal equivalence between LUPI and knowledge distillation, showing that teacher-student training on enriched data implements privileged supervision through soft-label transfer. Hinton et al.~\cite{hinton2015distilling} demonstrate that temperature-scaled KL divergence provides a rich inter-class relational signal beyond one-hot labels. Industrial applications confirm this paradigm: at Taobao, post-click behavioral signals—privileged at training but unavailable at serving—are distilled into click-through-rate predictors with significant accuracy gains~\cite{privileged2020xu}. In the security domain, KD has been applied to traffic classification primarily for model compression~\cite{knowledge2022huang,etkd2024pan}, not cross-modal generalization. To our knowledge, no prior WF work has formalized webpage resource structure as privileged information or exploited the \textit{training-rich / inference-poor} asymmetry to provide environment-agnostic supervision. ResAware fills this gap by treating the resource modality as a privileged teacher that transfers inter-class topology to a traffic-only student without expanding the online attacker's observational boundary.

\section{Conclusion}

This paper presents ResAware to address the performance degradation of WF models under environmental shift. By formalizing a \textit{training-rich / inference-poor} asymmetric threat model, ResAware uses stable application-layer resource sequences as privileged supervision to regularize traffic-only student models. Our findings show that internalizing resource-induced class topology allows the student to move beyond the observational limitations of the traffic modality, anchoring on a website's intrinsic identity rather than environment-specific traffic artifacts. Evaluated on a dataset spanning 5 months and 6 global vantage points, ResAware consistently improves the robustness of diverse WF architectures—including an 8.72\% F1-score gain for Var-CNN under 150-day temporal drift. With zero inference overhead and orthogonal compatibility with existing adaptation methods, ResAware provides a practical foundation for robust website fingerprinting in real-world deployments.

\bibliographystyle{ACM-Reference-Format}
\bibliography{refs}

\appendix

\section{Open Science}

Research artifacts have been de-identified for double-blind review and are hosted at: \url{https://github.com/aimafan123/ResAware}. The repository includes the full training and evaluation code for the ResAware distillation framework and teacher model, along with implementations of six WF backbones—AWF, DF, Var-CNN, Tik-Tok, RF, and CountMamba—reproduced from their original papers or official codebases.

The artifact suite also includes scripts for zero-shot evaluation across four drift scenarios (temporal, spatial, obfuscated proxy, and browser), open-world temporal drift testing, and few-shot adaptation supporting both supervised and Proteus-based unsupervised modes. Due to storage constraints and privacy considerations, we provide featurized versions of our cross-environment datasets. Automated pipelines for data processing, training, and evaluation are included to facilitate efficient reproduction of the core experimental results.

\section{Ethical Considerations}

This research follows the established ethical guidelines of the security community for network measurement and privacy analysis. Our practices are as follows.

\textbf{Data Collection and Privacy Protection.} 
All network traffic was generated by automated headless browsers on vantage points controlled by our team. We accessed only publicly indexable websites (a subset of the Tranco Top 100K~\cite{lepochat2019tranco}). Data collection complied with each site's \texttt{robots.txt} directives and terms of service. No real user browsing behavior, personally identifiable information (PII), or private communication content was involved at any stage. TLS session keys were extracted within our controlled client processes solely to reconstruct application-layer resource events and were neither retained beyond this purpose nor shared. Upon publication, we will release only featurized representations—packet direction and length sequences for traffic, and category and size sequences for resource loading. No raw payloads, URLs, IP addresses, or user-linked identifiers will be included, fully mitigating privacy risks while preserving research utility.

\textbf{Selection of Monitored Sites.} The 100 monitored sites were randomly sampled from the Tranco Top 100K list. To minimize exposure to politically sensitive content, we manually excluded sites identified by international human rights organizations as subject to mandatory censorship. The 83,645 unmonitored sites were also drawn from the Tranco rankings and do not target specific user groups or sensitive content.

\textbf{Dual-Use and Responsible Disclosure.} WF attack research is widely recognized as a prerequisite for improving privacy defenses: only by understanding and quantifying attacker capabilities can defenders design effective countermeasures. This paper focuses on the root causes of cross-environment WF failures rather than expanding online attacker capabilities. The core insight of ResAware—that unstable traffic-side supervision is the fundamental driver of generalization failure—provides direct guidance for defensive research. Defenders can introduce targeted perturbations at the resource-loading level—randomizing loading orders, injecting dummy requests, or diversifying resource type distributions—to undermine the stable resource-side inductive bias that ResAware exploits, complementing existing packet-level defenses. We will also release the large-scale paired traffic-resource dataset to lower the barrier for cross-environment WF research. Covering temporal, spatial, obfuscated proxy, and browser distribution shifts, this dataset represents one of the most comprehensive cross-environment WF benchmarks and will help the community evaluate attacks and defenses under unified protocols.

\section{ResAware Training and Deployment Protocol}
\label{sec:appendix_c}

Algorithm~\ref{alg:resaware} formalizes the three-stage training and deployment protocol of ResAware introduced in \S\ref{sec:design}. Stage~1 extracts two-channel privileged resource features from raw resource records and trains the resource-only teacher under hard-label supervision. Stage~2 freezes the teacher and distills its soft-target distributions into the traffic-only student through a weighted combination of classification loss and KL-divergence distillation loss. Stage~3 discards all resource-side components—the resource extractor, teacher model, cached soft labels, and distillation loss—before deployment, so that the deployed student operates on encrypted traffic alone with zero additional inference overhead.

\begin{algorithm}[t]
\caption{Resource-Privileged Distillation for Traffic-Only WF}
\label{alg:resaware}
\KwIn{Source-domain paired training set $\mathcal{D}_s=\{(x_i,R_i,y_i)\}_{i=1}^{n}$, where $R_i$ denotes raw resource records; resource teacher $T_{\theta_T}$; traffic student $S_{\theta_S}$; truncation length $N$; temperature $\tau$; distillation weight $\alpha$}
\KwOut{Deployed traffic-only student $S_{\theta_S}$}

\tcp{Stage 1: extract two-channel privileged resource features and train the teacher}
\ForEach{sample $(x_i,R_i,y_i)\in\mathcal{D}_s$}{
    $Z_i \leftarrow \text{SortByRequestOrder}(R_i)$\;
    $c_i \leftarrow \text{MapTypeToCategory}(Z_i)$ \tcp*[r]{categorical channel}
    $\tilde{s}_i \leftarrow \log(1+\text{PayloadBytes}(Z_i))$ \tcp*[r]{size channel}
    $x_i^* \leftarrow \text{Pad/Truncate}_N\big([(c_{i,1},\tilde{s}_{i,1}),\ldots,(c_{i,|Z_i|},\tilde{s}_{i,|Z_i|})]\big)$\;
}
Construct privileged set $\mathcal{D}_s^*=\{(x_i,x_i^*,y_i)\}_{i=1}^{n}$\;
\ForEach{mini-batch $\mathcal{B}^* \subset \{(x_i^*,y_i)\mid (x_i,x_i^*,y_i)\in\mathcal{D}_s^*\}$}{
    $z_T \leftarrow T_{\theta_T}(x^*)$\;
    $\mathcal{L}_T \leftarrow \mathrm{CE}(\sigma(z_T), y)$\;
    update $\theta_T$ by minimizing $\mathcal{L}_T$\;
}
Freeze $\theta_T$\;

\tcp{Stage 2: distill resource knowledge into the traffic-only student}
\ForEach{mini-batch $\mathcal{B} \subset \mathcal{D}_s^*$}{
    $z_T \leftarrow T_{\theta_T}(x^*)$ \tcp*[r]{privileged branch; no gradient to $T$}
    $z_S \leftarrow S_{\theta_S}(x)$ \tcp*[r]{deployable branch}
    $\mathcal{L}_{cls} \leftarrow \mathrm{CE}(\sigma(z_S), y)$\;
    $\mathcal{L}_{kd} \leftarrow \tau^2 D_{KL}(\sigma(z_T/\tau)\,\|\,\sigma(z_S/\tau))$\;
    $\mathcal{L}_{total} \leftarrow (1-\alpha)\mathcal{L}_{cls}+\alpha\mathcal{L}_{kd}$\;
    update $\theta_S$ by minimizing $\mathcal{L}_{total}$\;
}

\tcp{Stage 3: discard privileged components before deployment}
Discard resource extractor, $T_{\theta_T}$, cached soft labels, and distillation losses\;
\Return{$S_{\theta_S}$ for online inference on encrypted traffic $x$ only}\;
\end{algorithm}

\section{Dataset Construction and Collection Details}

\subsection{Methodology for Traffic-Resource Pairing}
\label{sec:a.1}
This section outlines the methodology for constructing trace-level traffic-resource pairs from a single controlled page visit. For each page load, we define a \textit{trace} as the aggregate network activity triggered by a complete page visit. For each crawler visit, we simultaneously capture raw encrypted traffic and the corresponding TLS session keys. In the offline phase, we reconstruct application-layer resource sequences and align them with the captured traffic traces. The detailed procedure for resource sequence recovery is provided in Algorithm~\ref{alg:resource_reconstruction}.

\begin{algorithm}[t]
\caption{Offline Privileged Resource Sequence Reconstruction}
\label{alg:resource_reconstruction}
\small
\KwIn{Encrypted packet sequence $x$, TLS session keys $K$}
\KwOut{Two-channel privileged resource sequence $x^*$}

\tcp{Recover application-layer records offline}
$x_{dec} \gets \text{DecryptTraffic}(x, K)$\;
$C \gets \emptyset$ \tcp*{Connection state map}
$Z \gets \emptyset$\;

\tcp{Group decrypted application frames into resource streams}
\ForEach{application frame $a \in x_{dec}$}{
    $f \gets \text{FlowTuple}(a)$\;
    $sid \gets \text{StreamID}(a)$\;

    \If{$sid \notin C[f]$}{
        $C[f][sid] \gets \text{NewStream}()$\;
    }
    $S \gets C[f][sid]$\;

    \If{$a \in \text{RequestHeaders}$}{
        $S.t_{req} \gets a.\text{time}$\;
    }
    \If{$a \in \text{ResponseHeaders}$}{
        $S.\text{type} \gets \text{InferResourceType}(a)$\;
    }
    \If{$a \in \text{DataFrame}$}{
        $S.\text{size} \gets S.\text{size} + a.\text{length}$\;
    }
}

\tcp{Keep complete streams and form a two-channel sequence}
\ForEach{stream $S \in C$}{
    \If{$S.\text{type}$ exists \textbf{and} $S.\text{size} > 0$}{
        $Z \gets Z \cup \{(S.t_{req}, S.\text{type}, S.\text{size})\}$\;
    }
}

$Z \gets \text{SortByRequestTime}(Z)$\;
$x^* \gets [(\text{type}_1,\text{size}_1),\ldots,(\text{type}_{|Z|},\text{size}_{|Z|})]$\;

\Return $x^*$\;
\end{algorithm}

\subsection{Dataset Information}
\label{sec:a.2}
To evaluate cross-environment robustness, we construct a dataset suite we term the \textit{ResAware Dataset Suite}. This suite is partitioned into multiple subsets, each isolating a distinct experimental factor: temporal evolution, spatial diversity, obfuscated proxy encapsulation, browser variation, and open-world background traffic.

To ensure environmental consistency across subsets, all vantage points run on Virtual Private Servers (VPS) hosted by Vultr\footnote{\url{https://www.vultr.com/}}. Each VPS uses an identical base configuration: Debian 13 OS, 1 vCPU, 2 GB RAM, 64 GB NVMe storage, and 2 TB bandwidth. During collection, two isolated Docker containers ran concurrently on each VPS to execute crawling tasks, providing a clean and reproducible environment. Each access targeted the site's homepage with a fixed 50-second capture window to cover initial page loads, asynchronous requests, and deferred resource loading. The automated browser scrolled the page three times at random intervals to trigger lazy-loaded images, scripts, and advertisement resources. After each visit, a screenshot was saved and a quality control (QC) pipeline filtered out failed visits, error pages, blank pages, and incomplete loads.

The collected subsets are described below:

\begin{itemize}
    \item \textbf{Train-Base:} The source-domain training set for temporal and spatial drift experiments. Collected on November 21, 2025 from 6 VPS in New York, US, using Chrome with standard HTTPS/TLS. It covers 100 monitored sites at 150 traces per site (15,000 paired traces total).

    \item \textbf{Open-World:} The unmonitored background pool for open-world evaluation, collected on November 21, 2025 from 6 VPS in New York, US with settings identical to \textit{Train-Base}. Starting from 100,000 Tranco Top 100K~\cite{lepochat2019tranco} candidate sites, we retain 83,645 after excluding monitored-set overlap and filtering failed or anomalous visits. Each site contributes one trace (83,645 total), used exclusively as the negative background pool.

    \item \textbf{Geo-Drift:} Used for spatial drift experiments. Collected on November 21, 2025 across five international vantage points—Japan (Tokyo), Singapore, South Africa (Johannesburg), Australia (Sydney), and Germany (Frankfurt)—using 10 VPS in total. All settings mirror \textit{Train-Base}. It covers the same 100 monitored sites at 25-30 traces per site per location (14,087 paired traces across five locations).

    \item \textbf{Time-Drift:} Used for temporal drift experiments, comprising five snapshots collected on December 21, 2025; January 20, 2026; February 19, 2026; March 21, 2026; and April 20, 2026—corresponding to 30, 60, 90, 120, and 150 days after \textit{Train-Base}. Each snapshot uses 2 VPS in New York, US, with 30 traces per site for the 100 monitored sites (15,000 paired traces across five snapshots).

    \item \textbf{Train-Base-2:} The source-domain training set for obfuscated proxy and browser drift experiments. Collected on March 21, 2026 from 6 VPS in New York, US with configurations identical to \textit{Train-Base}, covering 100 monitored sites at 150 traces per site (15,000 paired traces). Its temporal alignment with the obfuscated proxy and browser test sets controls for long-term temporal drift, so that observed performance differences are attributable primarily to protocol or browser variation.

    \item \textbf{Obfuscated-Proxy-Drift:} Used for obfuscated proxy drift experiments. Collected on March 21, 2026 using 12 client VPS in New York, US; all traffic was forwarded through Xray proxies. Two additional VPS served as Xray proxy servers, each handling three obfuscation protocols, with two client VPS assigned per protocol. Both clients and servers run Xray-core v26.1.23\footnote{\url{https://github.com/XTLS/Xray-core/releases/tag/v26.1.23}}. It covers 100 monitored sites at 30 traces per site per protocol (18,000 paired traces across six protocols).

    \item \textbf{Browser-Drift:} Used for browser drift experiments. Collected on March 21, 2026 using 4 VPS in New York, US (2 per browser: Edge and Firefox); all other settings match \textit{Train-Base-2}. It covers 100 monitored sites at 25-30 traces per browser per site (5,523 paired traces total).
\end{itemize}

\section{Complete Per-Environment Zero-Shot Results Across All Drift Scenarios}
\label{sec:all_result}

Table~\ref{tab:all_zero} reports the complete per-environment closed-world F1-score for all six backbones evaluated in \S\ref{sec:zero-shot}, complementing the aggregated results in Table~\ref{tab:drift_comparison}. Across 108 backbone-environment combinations, ResAware yields positive gains in 84 cases (77.78\%) and negative gains in 24 cases, indicating broad but not unconditional effectiveness. Negative cases mainly arise from low-capacity AWF under temporal/spatial/proxy/browser drift and from several protocol-induced proxy settings for DF/Tik-Tok, consistent with the applicability analysis in \S\ref{sec:applicability}.

\begin{table*}[t]
  \centering
  \caption{Complete per-environment zero-shot closed-world F1-score for all six backbones with (w/) and without (w/o) ResAware across all drift scenarios. Values are Mean $\pm$ SD.}
  \label{tab:all_zero}
\resizebox{\textwidth}{!}{
  \begin{tabular}{ll cc cc cc cc cc cc}
    \toprule
    \multirow{2}{*}{\textbf{Scenario}} & \multirow{2}{*}{\textbf{Target Env.}} & \multicolumn{2}{c}{\textbf{AWF}} & \multicolumn{2}{c}{\textbf{DF}} & \multicolumn{2}{c}{\textbf{RF}} & \multicolumn{2}{c}{\textbf{Tik-Tok}} & \multicolumn{2}{c}{\textbf{Var-CNN}} & \multicolumn{2}{c}{\textbf{CountMamba}} \\
    \cmidrule(lr){3-4} \cmidrule(lr){5-6} \cmidrule(lr){7-8} \cmidrule(lr){9-10} \cmidrule(lr){11-12} \cmidrule(lr){13-14}
    & & w/o & w/ & w/o & w/ & w/o & w/ & w/o & w/ & w/o & w/ & w/o & w/ \\
    \midrule

    \multirow{6}{*}{\shortstack[l]{Temporal\\Drift}}
    & Day 30  & 51.26 $\pm$ 9.16 & 51.04 $\pm$ 3.70 & 91.28 $\pm$ 0.36 & \textbf{91.72 $\pm$ 0.69} & 90.46 $\pm$ 0.95 & 89.99 $\pm$ 0.82 & 89.07 $\pm$ 0.66 & \textbf{90.39 $\pm$ 0.30} & 91.68 $\pm$ 0.80 & \textbf{93.95 $\pm$ 0.65} & 87.79 $\pm$ 1.32 & 87.06 $\pm$ 0.67 \\
    & Day 60  & 48.17 $\pm$ 8.36 & \textbf{49.00 $\pm$ 2.97} & 86.75 $\pm$ 1.24 & \textbf{87.25 $\pm$ 0.62} & 45.88 $\pm$ 1.94 & \textbf{47.71 $\pm$ 1.85} & 78.63 $\pm$ 1.89 & \textbf{81.45 $\pm$ 1.22} & 88.63 $\pm$ 1.78 & \textbf{93.38 $\pm$ 0.44} & 35.26 $\pm$ 3.31 & \textbf{36.09 $\pm$ 3.76} \\
    & Day 90  & 40.15 $\pm$ 6.10 & 39.58 $\pm$ 2.00 & 73.60 $\pm$ 0.67 & \textbf{76.46 $\pm$ 0.89} & 39.84 $\pm$ 1.89 & \textbf{42.37 $\pm$ 1.46} & 66.57 $\pm$ 1.13 & \textbf{69.72 $\pm$ 0.56} & 80.37 $\pm$ 1.95 & \textbf{86.37 $\pm$ 0.73} & 31.79 $\pm$ 2.03 & \textbf{32.81 $\pm$ 1.03} \\
    & Day 120 & 38.20 $\pm$ 5.35 & 37.89 $\pm$ 2.28 & 67.35 $\pm$ 0.92 & \textbf{71.38 $\pm$ 1.55} & 36.53 $\pm$ 2.04 & \textbf{38.73 $\pm$ 1.90} & 59.34 $\pm$ 1.14 & \textbf{62.67 $\pm$ 1.30} & 79.44 $\pm$ 1.99 & \textbf{87.60 $\pm$ 1.42} & 29.79 $\pm$ 2.89 & 29.38 $\pm$ 1.22 \\
    & Day 150 & 33.25 $\pm$ 5.23 & 32.25 $\pm$ 3.41 & 61.39 $\pm$ 1.11 & \textbf{65.79 $\pm$ 1.49} & 36.64 $\pm$ 2.46 & \textbf{38.27 $\pm$ 1.06} & 54.64 $\pm$ 0.84 & \textbf{57.67 $\pm$ 0.65} & 72.77 $\pm$ 1.63 & \textbf{81.49 $\pm$ 1.61} & 28.94 $\pm$ 2.34 & \textbf{29.16 $\pm$ 2.21} \\
    \cmidrule{2-14}
    & AVG     & 42.21 $\pm$ 6.63 & 41.95 $\pm$ 2.76 & 76.07 $\pm$ 0.56 & \textbf{78.52 $\pm$ 0.93} & 49.87 $\pm$ 1.69 & \textbf{51.41 $\pm$ 1.30} & 69.65 $\pm$ 1.01 & \textbf{72.38 $\pm$ 0.50} & 82.58 $\pm$ 1.54 & \textbf{88.56 $\pm$ 0.84} & 42.71 $\pm$ 2.33 & \textbf{42.90 $\pm$ 1.70} \\
    \midrule

    \multirow{6}{*}{\shortstack[l]{Spatial\\Drift}}
    & AU  & 53.83 $\pm$ 8.71 & 53.33 $\pm$ 4.08 & 87.04 $\pm$ 0.26 & \textbf{88.13 $\pm$ 0.45} & 79.01 $\pm$ 0.73 & \textbf{79.65 $\pm$ 0.72} & 85.44 $\pm$ 0.25 & \textbf{86.98 $\pm$ 0.29} & 84.74 $\pm$ 0.73 & \textbf{87.44 $\pm$ 0.54} & 74.87 $\pm$ 1.25 & \textbf{79.31 $\pm$ 0.95} \\
    & DE  & 40.80 $\pm$ 7.88 & 40.07 $\pm$ 3.16 & 81.99 $\pm$ 0.81 & \textbf{84.65 $\pm$ 1.11} & 77.21 $\pm$ 0.87 & \textbf{83.14 $\pm$ 0.62} & 80.87 $\pm$ 0.88 & \textbf{82.29 $\pm$ 0.31} & 81.99 $\pm$ 0.45 & \textbf{85.66 $\pm$ 1.07} & 77.05 $\pm$ 0.88 & 76.56 $\pm$ 0.18 \\
    & JP  & 50.95 $\pm$ 8.56 & 50.73 $\pm$ 3.98 & 84.50 $\pm$ 0.41 & \textbf{86.38 $\pm$ 0.22} & 78.80 $\pm$ 0.42 & \textbf{80.69 $\pm$ 0.50} & 83.05 $\pm$ 0.39 & \textbf{84.92 $\pm$ 0.24} & 83.21 $\pm$ 1.06 & \textbf{88.11 $\pm$ 0.68} & 74.14 $\pm$ 1.64 & \textbf{76.84 $\pm$ 0.37} \\
    & SG  & 51.64 $\pm$ 7.69 & 51.18 $\pm$ 2.84 & 86.84 $\pm$ 0.52 & \textbf{88.65 $\pm$ 0.33} & 79.79 $\pm$ 1.02 & \textbf{82.35 $\pm$ 0.34} & 83.97 $\pm$ 0.35 & \textbf{86.51 $\pm$ 0.58} & 84.71 $\pm$ 1.28 & \textbf{88.64 $\pm$ 0.21} & 75.44 $\pm$ 0.64 & \textbf{77.77 $\pm$ 0.21} \\
    & ZA  & 48.93 $\pm$ 5.55 & 48.50 $\pm$ 2.94 & 83.16 $\pm$ 1.02 & \textbf{85.37 $\pm$ 0.40} & 65.76 $\pm$ 1.95 & \textbf{67.23 $\pm$ 1.37} & 80.93 $\pm$ 0.75 & \textbf{84.79 $\pm$ 0.28} & 78.67 $\pm$ 0.54 & \textbf{84.97 $\pm$ 0.53} & 63.05 $\pm$ 1.48 & \textbf{69.69 $\pm$ 1.61} \\
    \cmidrule{2-14}
    & AVG & 49.23 $\pm$ 7.57 & 48.76 $\pm$ 3.36 & 84.71 $\pm$ 0.34 & \textbf{86.64 $\pm$ 0.28} & 76.11 $\pm$ 0.36 & \textbf{78.61 $\pm$ 0.38} & 82.85 $\pm$ 0.37 & \textbf{85.10 $\pm$ 0.22} & 82.66 $\pm$ 0.60 & \textbf{86.96 $\pm$ 0.40} & 72.91 $\pm$ 1.07 & \textbf{76.03 $\pm$ 0.59} \\
    \midrule

    \multirow{7}{*}{\shortstack[l]{Obfuscated\\Proxy\\Drift}}
    & Shadowsocks  & 12.09 $\pm$ 2.32 & \textbf{14.91 $\pm$ 2.16} & 43.94 $\pm$ 0.99 & \textbf{44.09 $\pm$ 0.91} & 59.32 $\pm$ 2.07 & \textbf{61.98 $\pm$ 3.23} & 45.90 $\pm$ 1.59 & 45.64 $\pm$ 1.20 & 40.21 $\pm$ 1.95 & \textbf{50.34 $\pm$ 2.88} & 60.84 $\pm$ 1.45 & \textbf{63.01 $\pm$ 0.16} \\
    & Trojan       & 13.02 $\pm$ 2.49 & \textbf{15.13 $\pm$ 1.95} & 45.26 $\pm$ 1.47 & \textbf{45.37 $\pm$ 0.97} & 63.27 $\pm$ 2.80 & \textbf{66.46 $\pm$ 2.56} & 46.77 $\pm$ 1.63 & 46.66 $\pm$ 1.46 & 42.07 $\pm$ 1.36 & \textbf{53.25 $\pm$ 2.49} & 64.47 $\pm$ 2.05 & \textbf{64.58 $\pm$ 0.03} \\
    & VLESS-XTLS-Vision        & 18.48 $\pm$ 0.88 & 16.77 $\pm$ 0.54 & 41.33 $\pm$ 1.14 & 40.05 $\pm$ 0.69 & 48.30 $\pm$ 1.97 & \textbf{51.58 $\pm$ 1.99} & 35.94 $\pm$ 1.84 & \textbf{37.12 $\pm$ 0.55} & 29.33 $\pm$ 2.65 & \textbf{36.25 $\pm$ 3.16} & 46.89 $\pm$ 1.88 & \textbf{47.72 $\pm$ 0.54} \\
    & VMess-TLS    & 15.32 $\pm$ 2.82 & \textbf{18.09 $\pm$ 2.10} & 48.90 $\pm$ 1.38 & 47.38 $\pm$ 1.37 & 64.33 $\pm$ 2.17 & \textbf{67.53 $\pm$ 22.9} & 45.16 $\pm$ 3.11 & \textbf{45.47 $\pm$ 0.88} & 45.44 $\pm$ 1.82 & \textbf{55.73 $\pm$ 2.13} & 63.91 $\pm$ 0.46 & \textbf{65.41 $\pm$ 1.32} \\
    & VMess        & 25.03 $\pm$ 0.74 & 22.94 $\pm$ 2.04 & 57.68 $\pm$ 1.28 & 55.11 $\pm$ 0.66 & 72.02 $\pm$ 2.21 & \textbf{76.90 $\pm$ 2.63} & 47.76 $\pm$ 4.18 & \textbf{48.69 $\pm$ 1.52} & 41.46 $\pm$ 3.07 & \textbf{52.21 $\pm$ 3.11} & 65.34 $\pm$ 0.62 & \textbf{66.37 $\pm$ 1.23} \\
    & VMess-WS-TLS     & 21.25 $\pm$ 0.94 & 20.32 $\pm$ 1.73 & 52.82 $\pm$ 1.64 & 51.70 $\pm$ 0.49 & 69.94 $\pm$ 2.13 & \textbf{75.99 $\pm$ 2.29} & 45.57 $\pm$ 3.73 & \textbf{45.71 $\pm$ 1.60} & 30.30 $\pm$ 5.81 & \textbf{34.85 $\pm$ 3.15} & 65.82 $\pm$ 0.51 & \textbf{67.89 $\pm$ 0.83} \\
    \cmidrule{2-14}
    & AVG          & 17.53 $\pm$ 1.34 & \textbf{18.03 $\pm$ 1.45} & 48.32 $\pm$ 0.77 & 47.28 $\pm$ 0.73 & 62.86 $\pm$ 2.12 & \textbf{66.74 $\pm$ 2.36} & 44.52 $\pm$ 2.34 & \textbf{44.88 $\pm$ 0.35} & 38.14 $\pm$ 1.81 & \textbf{47.10 $\pm$ 2.46} & 61.21 $\pm$ 1.04 & \textbf{62.50 $\pm$ 0.44} \\
    \midrule

    \multirow{3}{*}{\shortstack[l]{Browser\\Drift}}
    & Edge    & 10.23 $\pm$ 1.88 & 09.98 $\pm$ 1.09 & 07.66 $\pm$ 0.69 & \textbf{12.18 $\pm$ 0.42} & 27.08 $\pm$ 1.57 & \textbf{32.60 $\pm$ 0.41} & 09.18 $\pm$ 0.52 & \textbf{11.80 $\pm$ 0.85} & 26.16 $\pm$ 2.11 & \textbf{33.64 $\pm$ 0.98} & 12.80 $\pm$ 1.22 & \textbf{16.37 $\pm$ 2.28} \\
    & Firefox & 01.60 $\pm$ 0.26 & \textbf{02.15 $\pm$ 0.29} & 00.49 $\pm$ 0.10 & \textbf{01.13 $\pm$ 0.22} & 09.22 $\pm$ 0.64 & \textbf{13.07 $\pm$ 1.37} & 00.40 $\pm$ 0.23 & 00.30 $\pm$ 0.18 & 08.31 $\pm$ 0.68 & \textbf{09.27 $\pm$ 1.59} & 01.42 $\pm$ 0.45 & \textbf{02.63 $\pm$ 0.24} \\
    \cmidrule{2-14}
    & AVG     & 05.91 $\pm$ 0.87 & \textbf{06.06 $\pm$ 0.69} & 04.07 $\pm$ 0.32 & \textbf{06.66 $\pm$ 0.21} & 18.15 $\pm$ 1.04 & \textbf{22.83 $\pm$ 0.82} & 04.79 $\pm$ 0.30 & \textbf{06.05 $\pm$ 0.47} & 17.24 $\pm$ 1.20 & \textbf{21.45 $\pm$ 0.72} & 07.11 $\pm$ 0.51 & \textbf{09.50 $\pm$ 1.02} \\
    \bottomrule
  \end{tabular}
  }
\end{table*}
\section{Sensitivity Analysis of \texorpdfstring{$\alpha$}{alpha}}
\label{sec:appendix_alpha}

\begin{figure*}[t]
    \centering
    \includegraphics[width=\textwidth]{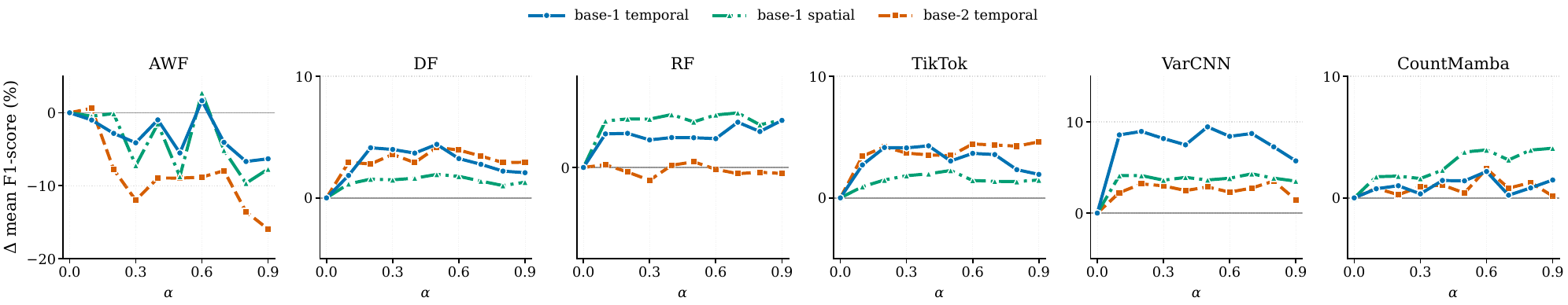}
    \caption{Performance gain $\Delta$ (\%) over the $\alpha=0$ baseline as a function of distillation weight $\alpha$ for six backbones. The best-performing $\alpha$ range remains largely stable for each backbone, indicating that the distillation weight is mainly coupled to student capacity rather than to a particular source training window.}
    \label{fig:alpha_ab}
\end{figure*}

This appendix provides the complete sensitivity analysis of the distillation weight $\alpha$. Our goal is to understand how strongly resource-privileged supervision should be injected into the traffic-only student, and whether this choice is tied to a particular target environment or instead reflects an intrinsic property of the student backbone.

In the joint objective
\begin{equation}
    \mathcal{L}_{total} = (1-\alpha)\mathcal{L}_{cls} + \alpha\mathcal{L}_{kd},
\end{equation}
the weight $\alpha \in [0,1]$ controls how much optimization pressure is assigned to the resource-induced \textit{inter-class topology}, relative to hard-label discrimination. When $\alpha$ is too small, the student receives little structural supervision from the resource teacher and largely degenerates to ordinary traffic-only ERM. When $\alpha$ is too large, the student may be forced to fit soft topological constraints that exceed the representational capacity of its traffic-side feature space.

To characterize this trade-off, we perform a full $\alpha$ scan over all six student backbones under temporal drift from two independent training datasets. We additionally verify the same trend under spatial drift from the source training dataset. Figure~\ref{fig:alpha_ab} reports the temporal-drift scan and shows that the best single $\alpha$ may shift slightly across training and testing datasets, but each backbone exhibits a stable best-performing range. This range is primarily governed by student capacity rather than the specific training window. DF, Tik-Tok, and RF benefit most from moderate distillation weights, whereas Var-CNN and CountMamba benefit from medium-to-high distillation weights; AWF is the most sensitive and can degrade when the distillation term dominates.

This pattern indicates that $\alpha$ should be interpreted as a capacity-matching parameter on the student side. The resource teacher's inter-class topology must ultimately be compressed into a \textit{traffic-only} representation space. A higher-capacity student can absorb this structural prior while preserving hard-label decision boundaries; a lower-capacity student has a smaller representation budget and is more prone to objective interference between $\mathcal{L}_{cls}$ and $\mathcal{L}_{kd}$.

\noindent \textbf{Var-CNN Capacity Scaling.}
To further isolate the effect of model capacity, we keep the residual topology of Var-CNN fixed and scale only the channel width. We then repeat the $\alpha$ scan under 150-day temporal drift. This controlled experiment removes architectural differences and concentrates the comparison on student capacity itself. Table~\ref{tab:capacity_scaling} summarizes the best $\alpha$ range for each width, where the best range denotes weights that remain close to the capacity-specific optimum and clearly outperform the $\alpha=0$ baseline.

\begin{table}[t]
\centering
\caption{Capacity scaling analysis for Var-CNN under 150-day temporal drift. The best $\alpha$ range denotes distillation weights that remain close to the best result for each width and outperform the $\alpha=0$ baseline.}
\label{tab:capacity_scaling}
\resizebox{\columnwidth}{!}{
\begin{tabular}{lcccc}
\toprule
\textbf{Var-CNN Width} & \textbf{$\alpha=0$} & \textbf{Best $\alpha$ Range} & \textbf{F1 in Best Range} & \textbf{Max Gain} \\
\midrule
$1\times$     & 72.77 & 0.1--0.7 & 80.25--82.22 & +9.45 pp \\
$0.5\times$   & 74.36 & 0.1--0.6 & 80.31--80.94 & +6.58 pp \\
$0.25\times$  & 72.18 & 0.2--0.4 & 76.82--78.35 & +6.17 pp \\
$0.125\times$ & 70.35 & 0.1--0.3 & 73.14--74.95 & +4.60 pp \\
\bottomrule
\end{tabular}
}
\end{table}

Table~\ref{tab:capacity_scaling} shows that capacity controls both the ceiling of distillation gains and the best-performing range of $\alpha$. The full-width Var-CNN has the widest best range: for $\alpha=0.1$--$0.7$, it maintains 80.25\%--82.22\% macro-F1 and reaches a maximum gain of 9.45 percentage points. Reducing the width to $0.5\times$ still preserves a broad best range of $\alpha=0.1$--$0.6$, but the maximum gain drops to 6.58 percentage points. Further reducing the width to $0.25\times$ and $0.125\times$ narrows the best ranges to $0.2$--$0.4$ and $0.1$--$0.3$, respectively, while the maximum gains decrease to 6.17 and 4.60 percentage points. Under the same teacher, training data, and residual topology, smaller students therefore convert less privileged resource supervision into peak robustness gains and exhibit narrower near-optimal $\alpha$ ranges.

This result does not imply that low-capacity students cannot benefit from resource supervision. Rather, they can absorb only a limited strength of teacher topology. For high-capacity students, the KD term mainly acts as a structural regularizer that reshapes decision boundaries without overwhelming hard-label discrimination. For low-capacity students, an overly large $\alpha$ allocates too much of the representation budget to matching the teacher distribution, turning the KD term from useful regularization into an optimization constraint beyond the student's capacity.

\noindent \textbf{Takeaways.}
The distillation weight $\alpha$ should be viewed as the strength of resource-privileged supervision matched to student capacity. Architectural differences affect the exact optimum, but the underlying mechanism is whether the student has sufficient representation budget to internalize the resource teacher's inter-class topology. In practice, we tune $\alpha$ once per backbone on the source-domain validation set and keep it fixed across all target environments; the reported robustness gains do not rely on target-domain retuning. This analysis also clarifies the applicability boundary of ResAware: when the student has adequate capacity and the correspondence between resource structure and traffic observations remains stable, moderate or large $\alpha$ can substantially improve cross-environment generalization; when student capacity is limited, $\alpha$ should be reduced to avoid over-distillation; when drift disrupts the cross-modal correspondence itself, $\alpha$ should be further reduced or the model should fall back to traffic-only training.

% Content moved to the main body: \S\ref{sec:applicability}

\end{document}